\documentclass{article} 
\usepackage{iclr2025_conference,times}

\usepackage{hyperref}
\usepackage{url}
\usepackage{graphicx}
\usepackage{algorithm}
\usepackage{algpseudocode}
\usepackage{multirow}
\usepackage{booktabs}
\usepackage{amssymb}

\usepackage{scalerel}
\newcommand{\fire}[0]{\scalerel*{\includegraphics{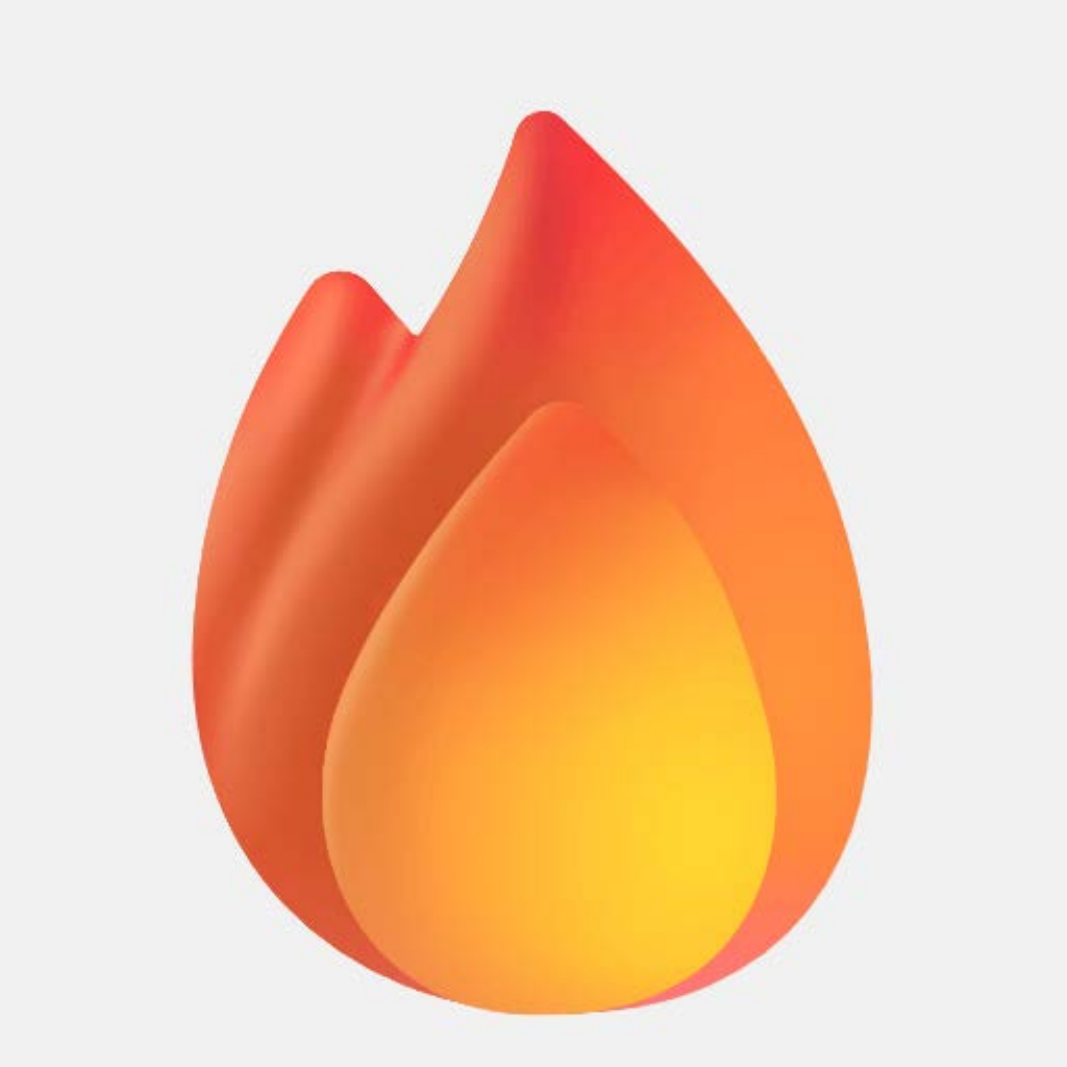}}{F}}
\newcommand{\snow}[0]{\scalerel*{\includegraphics{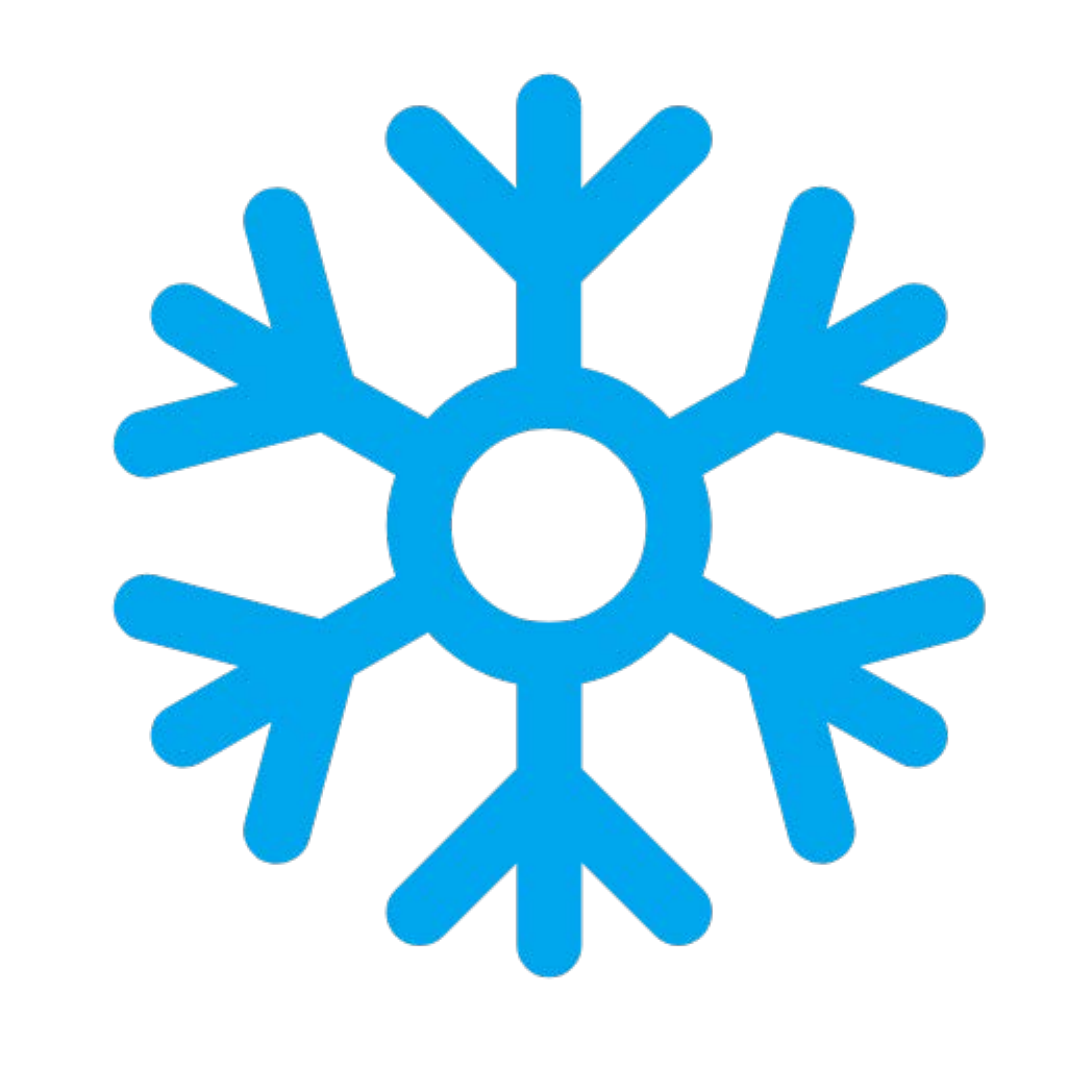}}{F}}

\title{Understanding Multimodal Hallucination with Parameter-Free Representation Alignment}

\author{Yueqian Wang \& Jianxin Liang \& Huishuai Zhang$^\ast$ \\
Wangxuan Institute of Computer Technology, Peking University\\
\texttt{\{wangyueqian,liangjx,zhanghuishuai\}@pku.edu.cn} \\
\And
Yuxuan Wang\\
Beijing Institute for \\
General Artificial Intelligence \\
\texttt{wangyuxuan1@bigai.ai} \\
\And
Dongyan Zhao\thanks{Corresponding Authors}\\
Wangxuan Institute of Computer Technology, Peking University\\
National Key Laboratory of General Artificial Intelligence\\
\texttt{zhaodongyan@pku.edu.cn}
}

\iclrfinalcopy 
\begin{document}

\maketitle

\begin{abstract}
Hallucination is a common issue in Multimodal Large Language Models (MLLMs), yet the underlying principles remain poorly understood. In this paper, we investigate which components of MLLMs contribute to object hallucinations. To analyze image representations while completely avoiding the influence of all other factors other than the image representation itself, we propose a \textbf{p}arametric-\textbf{f}ree \textbf{r}epresentation \textbf{a}lignment \textbf{m}etric (Pfram) that can measure the similarities between any two representation systems without requiring additional training parameters.
Notably, Pfram can also assess the alignment of a neural representation system with the human representation system, represented by ground-truth annotations of images. By evaluating the alignment with object annotations, we demonstrate that this metric shows strong and consistent correlations with object hallucination across a wide range of state-of-the-art MLLMs, spanning various model architectures and sizes. Furthermore, using this metric, we explore other key issues related to image representations in MLLMs, such as the role of different modules, the impact of textual instructions, and potential improvements including the use of alternative visual encoders. Our code is available at: \href{https://github.com/yellow-binary-tree/Pfram}{https://github.com/yellow-binary-tree/Pfram}.
\end{abstract}

\section{Introduction}
Multimodal Large Language Models (MLLMs) have been rapidly advancing in recent days \cite{Dai2023InstructBLIPTG,Liu2023VisualIT,Liu2023ImprovedBW,Zhang2023InternLMXComposerAV,Dong2024InternLMXComposer2MF,Bai2023QwenVLAF}. By connecting visual encoders with Large Language Models (LLMs), MLLMs integrate the capability to understand images within LLMs, showcasing their value in various applications such as visual question answering, image captioning, and multimodal conversation.

Despite their achievements and potential, MLLMs have a fundamental shortcoming that remains unaddressed: they struggle to accurately understand details in images, often generating responses that contradict the image content, a phenomenon known as ``hallucination''. Although many works have constructed benchmarks to evaluate hallucination in MLLMs in an end-to-end manner and have explored various approaches to mitigate hallucination, research on the underlying principles of hallucinations during the image understanding process in MLLMs is relatively scarce.

\begin{figure}
    \centering
    
    \begin{minipage}{0.48\textwidth}
    \centering
    \includegraphics[width=0.7\linewidth]{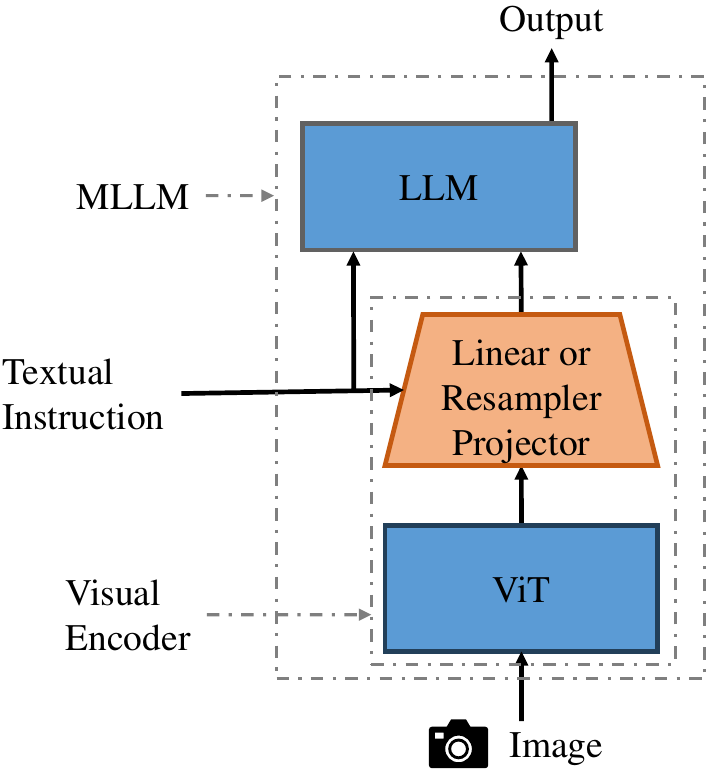}
    \caption{Terms of different parts of MLLMs used in this paper.}
    \label{fig:terms}
    \end{minipage}\quad
    \begin{minipage}{0.48\textwidth}
    \centering
    \includegraphics[width=0.9\linewidth]{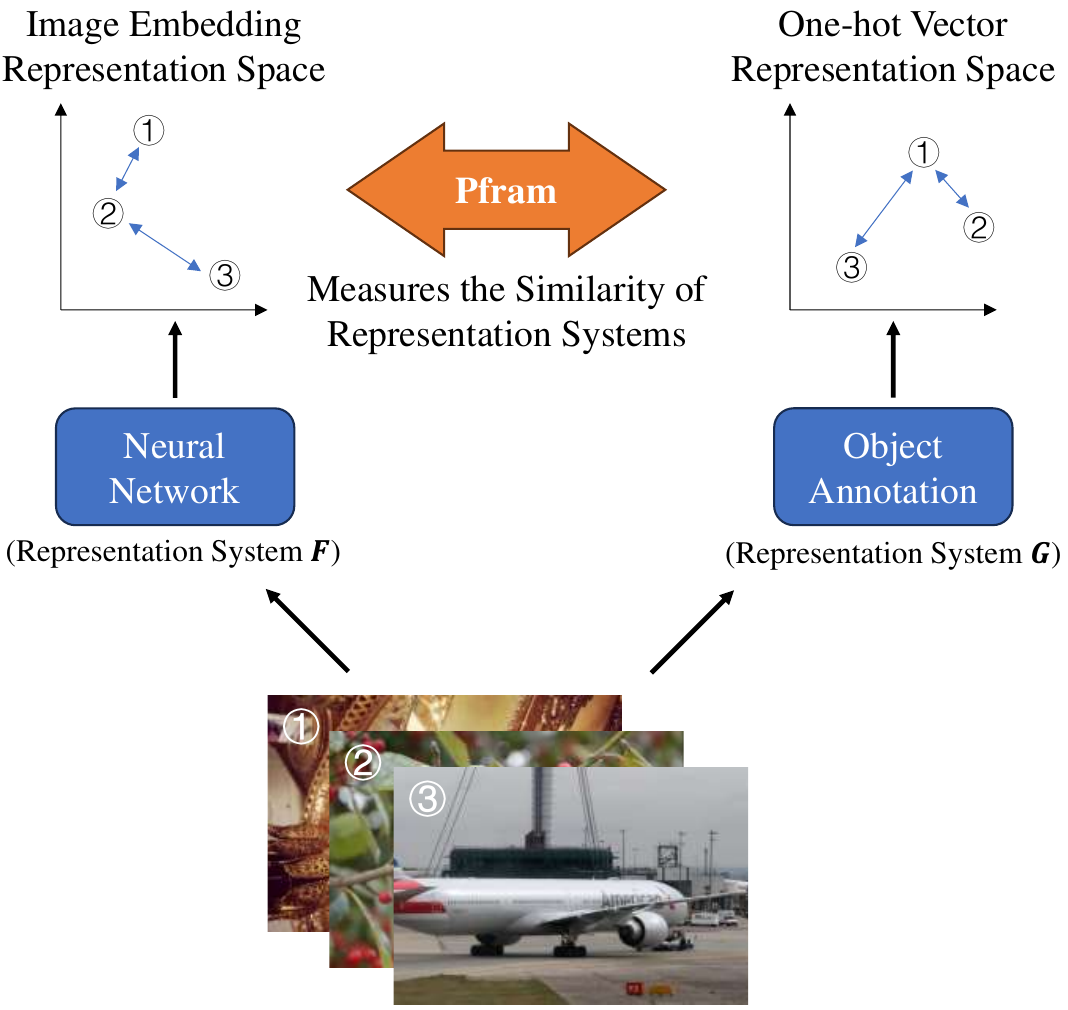}
    \caption{A demonstration of the Pfram metric.}
    \label{fig:overview}
    \end{minipage}
\end{figure}

In this paper, we focus on a critical question: which part of an MLLM's functioning, e.g., the visual encoder or the LLM as illustrated in Figure \ref{fig:terms}, contributes more to the hallucination problem?  Alternatively, we can frame the question as follows: Does the visual encoder fail to capture image details, leading to low-quality image representations, or is the issue rooted in the LLM's ability to understand multi-modal context? Existing visual hallucination benchmarks evaluate the entire MLLM as a whole in an (image + text)-to-text manner, making it challenging to isolate the effects of different components. Therefore, we need a method to evaluate image representations independently. 

To clarify our discussion, we define some key concepts used throughout this paper:

\begin{itemize}
    \item An \textbf{image representation} $r$ is a depiction of an image, which can be continuous, such as features generated by neural networks, or discrete, like captions or lists of ground-truth object labels. Our only requirement for an image representation is that the similarity between representations of any pair of images must be computable.   
    \item A \textbf{representation system} is a function $\mathcal{F}: I \rightarrow r$ that transforms an image $I$ into a representation $r$. This function can be an image encoder, an object detector, or a human annotator.
\end{itemize}

In this paper, we propose a \textbf{p}arametric-\textbf{f}ree \textbf{r}epresentation \textbf{a}lignment \textbf{m}etric (Pfram) that can measure the similarities between any two representation systems without requiring additional training parameters. Interestingly, we can employ Pfram to evaluate the performance of each component of an MLLM by comparing its image representations with the human representation system, i.e., the ground-truth annotations, with regard to certain information of some aspect. If the quantity of certain information in these representations strongly correlates with the hallucination performance of the entire MLLM, while other factors do not, we can conclude that the image understanding process is the primary factor influencing hallucination.  Therefore, this metric is well-suited for isolating the effects of image understanding from other components of an MLLM concerning hallucination.

Specifically, in this work, we particularly consider object hallucination, an important and feasible research direction for MLLMs, as it can be automatically and accurately measured using discriminative benchmarks such as \cite{Li2023EvaluatingOH}.  Suppose that human annotation representations perfectly preserve object information and hence we use human annotations of objects as the ground-truth representation system.
For any representation system, e.g., visual encoders of MLLMs, the more ``similar'' it is to the ground-truth representation system, the better it is at preserving object information. This ``similarity'' is not evaluated using single-example metrics, which compare representations of two systems for the same image, because the similarity between image representations generated by neural networks and object labels annotated by humans cannot be directly computed in a parameter-free manner. Instead, we employ ranking-based or neighbor-based metrics: for each image, we rank the similarities of all other images to it (or retrieve the k-nearest neighboring images) using both the representation system to evaluate and the object label annotations. The closer the two rankings (or the larger the intersection of the two k-nearest neighbor sets) are, the better the representation system is at preserving object information in images. 

The proposed Pfram metric enjoys several merits: (1)  it is applicable to all hidden layers' representations across all types of encoders, as long as the similarity of these hidden representations can be computed;  (2) It is completely training-free and does not introduce any new parameters, which is highly efficient and enables the applicability in various scenarios. (3) It is highly flexible for customization, able to measure the similarity between neural network representation and ground-truth annotations.

We note that comparing similarities between two neural networks' representations has been widely studied for insights into learning dynamics, robustness, and generalizability of neural networks \cite{Klabunde2023SimilarityON}. Our Pfram approach is inspired by these methods but replaces one of the neural networks with ground-truth annotations, providing new capabilities.

Our contribution can be summarized as follows.
\begin{itemize}
    \item We propose a parameter-free representation alignment metric (Pfram), which can be used to evaluate any specified aspects of image representations in a completely parameter-free manner.
    \item By applying the Pfram to investigate object hallucination across a wide range of recent state-of-the-art MLLMs, we demonstrate that object hallucination has a strong and consistent correlation with object information contained in image representations, which can be quantified by Pfram. In contrast, other factors, such as model size, model structure or alignment to image descriptions, do not show such a strong correlation.
    \item Using Pfram, we diagnose the effect of different modules of MLLM related to multimodal hallunication,  including the role of different modules in the image processing of MLLMs, the effect of using textual instructions, and potential improvements such as alternative Vision Transformers (ViTs) for MLLMs. 
\end{itemize}

\section{Related Works}
\subsection{Hallucination of MLLMs}
Current benchmarks of evaluating MLLMs' hallucination focus on two main aspects: hallucination discrimination methods \cite{Li2023EvaluatingOH,Hu2023CIEMCI,Lovenia2023NegativeOP,Wang2024VideoHallucerEI} that ask MLLM whether objects exist in the image in a visual question answering format, or non-hallucinatory generation methods \cite{Rohrbach2018ObjectHI,Liu2023MitigatingHI,Gunjal2023DetectingAP,Jing2023FAITHSCOREEH,Wang2023EvaluationAA,Sun2023AligningLM,Wang2023AnLM} that evaluate how much hallucinatory content exist in image descriptions generated by MLLMs.

There are also some works that study mitigating hallucinations from different perspectives, including enriching and balancing training data \cite{Hu2023CIEMCI,You2023FerretRA,Gunjal2023DetectingAP,Zhai2023HallESwitchRA}, improving alignment objectives \cite{Sun2023AligningLM,Yu2023RLHFVTT,Jiang2023HallucinationAC}, decoding strategies \cite{Huang2023OPERAAH,Leng2023MitigatingOH} and post-processing \cite{Yin2023WoodpeckerHC,Zhou2023AnalyzingAM}. 

Some works tried to explain the principles behind hallucinations of MLLMs. OPERA \cite{Huang2023OPERAAH} analyses the attention map of MLLMs and finds that the ``Aggregation pattern'' leads to hallucination of current MLLMs. \citet{Zhou2023AnalyzingAM} show that co-occurrence patterns, uncertainty, and token position are three primary factors contributing to object hallucination. HallE-Switch \cite{Zhai2023HallESwitchRA} claims that training with language descriptions that include details at a finer granularity than what the vision module can ground or verify induces hallucinations. Compared to these works, our work is the first one that studies hallucinations through image representations, and covers the widest range of state-of-the-art open-source MLLMs.

We refer readers to \citet{Bai2024HallucinationOM} for a recent comprehensive survey about hallucination in MLLMs.

\subsection{Similarity Metrics of Representations} 
\label{sec:related_similarity}
There are many measures to calculate the similarity of different models according to their representations. These measures can be roughly divided into canonical correlation analysis \cite{Raghu2017SVCCASV,Morcos2018InsightsOR}, representation alignment \cite{Ding2021GroundingRS,Williams2021GeneralizedSM,Li2015ConvergentLD}, similarity matrix \cite{Shahbazi2020UsingDO,Kriegeskorte2008RepresentationalSA,Szekely2007MeasuringAT}, neighbours \cite{Hryniowski2020InterlayerIS,Wang2020TowardsUT,Huh2024ThePR} and statistic \cite{Wang2020TowardsUT,Wang2020UnderstandingTB,Gwilliam2022BeyondSV} based methods. 

To the best of our knowledge, all previous works have focused on measuring similarity among the same types of representation systems, such as neural networks. This work is the first to treat ground-truth annotations as a representation system, evaluating the representation abilities of neural networks by calculating their similarity with these annotations.

\section{Introduction of The Pfram Metric}
The Pfram metric is formally defined as:
\begin{equation}
\mathrm{Pfram}(\mathcal{F}; \mathcal{G}| \phi, \mathcal{I}) :=  \frac{1}{n}\sum_{i=1}^n \phi(\mathcal{F}; \mathcal{G}| I_i, \mathcal{I}). 
\end{equation}
We note that the Pfram metric has four variables, i.e., $\mathcal{F}$ is the representation system we want to measure, $\mathcal{G}$ is the ground-truth representation system that captures concerning information, $\mathcal{I}=\{I_1, I_2, \cdots, I_n\}$ is a set of $n$ images, and $\phi$ is an system-level similarity metric that calculates the similarity of $\mathcal{F}$ and $\mathcal{G}$ for a given ``anchor image'' $I_i$ and the reference set of images  $\mathcal{I}$. 

Specifically, our choice of $\phi$ ensures that $\phi(\mathcal{F}; \mathcal{G} |I_i, \mathcal{I}) \in [0, 1]$ and hence $\mathrm{Pfram}(\mathcal{F}; \mathcal{G}| \phi, \mathcal{I}) \in [0,1]$ where the larger value of Pfram means the more similarity between $\mathcal{F}$ and $\mathcal{G}$.

In this paper, $\mathcal{F}$ is usually a layer of a pre-trained model, thus $\mathcal{F}$ can be further denoted as $\mathcal{F} := (\mathcal{M}, l)$, where $\mathcal{M}$ is the model, e.g., LLaVA \cite{Liu2023VisualIT}, and $l$ is the layer of $\mathcal{M}$ we want to measure, e.g., $l=0$ denotes the input layer of the LLM component of the MLLM.

As the aspect of image representation we want to measure is object information, we need to choose a $\mathcal{G}$ that reflects object information in images. We use the human annotation of objects (such as ``person'', ``cup'' or ``grass'') that appear in an image and denote it as $\mathcal{G}_{obj}:\mathcal{G}_{obj}(I) \in \{0,1\}^{m}$ where $m$ is the size of the object vocabulary, so the $j^{\mathrm{th}}$ item of $\mathcal{G}_{obj}(I)$ denotes whether the $j^{\mathrm{th}}$ object in the object vocabulary appears in image $I$.

To demonstrate the necessity of using object annotations for interpreting MLLMs' object hallucinations, or in other words, models that perform better at object hallucination really have better image representations in terms of object information instead of a general understanding of images, we also use another $\mathcal{G}$ that represents an image with its brief descriptions. We prompt GPT-4o with
instruction “Provide a brief introduction of this image with
one sentence.” to get a one-sentence brief description of the
image, mimicking the case that the description is brief in
most corpora, and use a sentence transformer \cite{Reimers2019SentenceBERTSE} fine-tuned from MPNet-base \cite{Song2020MPNetMA} to acquire a representation of the description. We denote this system as $\mathcal{G}_{des}:\mathcal{G}_{des}(I)=\mathrm{MPNet}(\mathrm{GPT4o(I)}) \in \mathbb{R}^h$, where $h$ is the hidden size of MPNet.

For the system-level similarity metric $\phi$, we use
a neighbor-based metric, i.e., Mutual k-NN \cite{Hryniowski2020InterlayerIS}, and a ranking-based metric, i.e., Normalized Discounted Cumulative Gain (NDCG) \cite{Jrvelin2000IREM,Jrvelin2002CumulatedGE}. We denote them as $\phi_\mathrm{kNN}$ and $\phi_\mathrm{NDCG}$, respectively.
Mutual k-NN measures similarity based on nearest neighbors in the respective representation spaces. A higher overlap of the two sets of nearest neighboring images  for one anchor image indicates greater similarity between the two representation systems.
NDCG is commonly used to evaluate retrieval systems based on their rankings and the ground-truth relevance (gains) of a list of items.
The main difference between NDCG@k and Mutual k-NN is that NDCG@k considers the order of images in the ranked list, giving higher weights to those more similar to the anchor image, while Mutual k-NN does not take the order information into account. Formal descriptions of the two metrics are listed in the appendix.

To calculate the system-level similarity metric either Mutual k-NN or NCDG@k, an image-level similarity between two images, i.e., an anchor image $I_{anc}$ and an any other image from the dataset, which is referred to as a ``reference image'' $I_{ref}$,  for a given representation system $\mathcal{G}$ or $\mathcal{F}$ should firstly be defined. For $\mathcal{G}_{obj}$, the number of overlapping objects, i.e., $\langle \mathcal{G}_{obj}(I_{anc}), \mathcal{G}_{obj}(I_{ref})\rangle$, is used as the image-level similarity. For $\mathcal{G}_{des}$, the cosine similarity between two description representations, i.e. $\frac{\langle \mathcal{G}_{des}(I_{anc}), \mathcal{G}_{des}(I_{ref}) \rangle}{\Vert \mathcal{G}_{des}(I_{anc}) \Vert_2 \Vert \mathcal{G}_{des}(I_{ref}) \Vert_2}$, is used to measure the image-level similarity.

For $\mathcal{F}$, it is important to note that transformers generate image features consisting of multiple vectors. For example, a Vision Transformer (ViT) produces a vector for each image patch, resulting in a list of \texttt{num\_patches} vectors. Similarly, Qformers or Resamplers, which are common structures used for image encoding in MLLMs \cite{Li2023BLIP2BL,Dai2023InstructBLIPTG,Bai2023QwenVLAF}, produce a list of \texttt{num\_query} vectors (usually 32 or 64) for each image. Different vectors in this list may correspond to different spatial positions within the image, meaning the same object appearing in different regions may result in features related to the object appearing in different feature vectors. To address the impact of an object appearing in different regions across images, for each vector in $\mathcal{F}(I_{anc})$, we find its largest cosine similarity with all vectors of $\mathcal{F}(I_{ref})$ and use the mean of these largest values to be the final similarity. This process is formally described in Algorithm \ref{alg:cossim}.

\begin{algorithm}[t]
\caption{Similarity Between 2 Image Representations} \label{alg:cossim}
\begin{algorithmic}
\Require 
    \State A representation system $\mathcal{F}$;
    \State An anchor image ${I}_{anc}$ and a reference image ${I}_{ref}$.
\Ensure A similarity score $sim$.

\State $H_{src} = \mathcal{F}(I_{anc}) = [r_{anc}^1, r_{anc}^2, \ldots, r_{anc}^N] \in \mathbb{R}^{N \times d}$,
\State $H_{ref} = \mathcal{F}(I_{ref}) = [r_{ref}^1, r_{ref}^2, \ldots, r_{ref}^N] \in \mathbb{R}^{N \times d}$,
\State $sim\_scores = [], i = 1$
\While{$i \leq N$}
    \State $s = \max(\cos(r_{anc}^i, r_{ref}^1), \ldots, \cos(r_{anc}^i, r_{ref}^N))$
    \State Append $s$ into $sim\_scores$
    \State{$i = i + 1$}
\EndWhile
\State $sim = \frac{\sum(sim\_scores)}{N}$
\end{algorithmic}
\end{algorithm}

\section{The Correlation between Pfram and Object Hallucination}

In this section, we use Pfram to assess how well the representations from an MLLM's visual encoder align with human object annotations. Our findings reveal a strong correlation between Pfram$(\mathcal{F}; \mathcal{G}_{obj})$ and the MLLM's tendency to hallucinate objects. We begin by introducing the models and datasets used in our evaluation.

\subsection{Implementation Details}
\paragraph{Models and datasets.}
We select 10 state-of-the-art open-source MLLMs, representing a variety of model architectures, parameter counts, training datasets, and fine-tuning methods. Please refer to the appendix for the descriptions of these models.

The image dataset $\mathcal{I}$ used in our evaluation adhere to the following criteria: (1) They are not included in the training data of the MLLMs we are evaluating, ensuring a fair comparison and preventing any data leakage issues; (2) They have manually annotated object labels with a controlled vocabulary, which serves as the representation system $\mathcal{G}_{obj}$; and (3) They contain at least a minimum number of object classes per image (for example, at least 5 object classes per image) to enhance the distinguishability of the intersection sizes as the image-level similarity for $\mathcal{G}$. Based on these requirements, we select two image datasets: AMBER \cite{Wang2023AnLM} and the validation set of OpenImages V7 (OIv7 for short) \cite{OpenImages}. Since there are only 638 images with at least 5 objects in AMBER, we use all of them ($n=638$) for our experiments. Due to the larger size of OIv7, we randomly sample $n=1600$ images with at least 5 object classes for an experiment and conduct three experiments with different random samples.

Following POPE \cite{Li2023EvaluatingOH}, we randomly sample 500 images for each experiment, generating 3,000 yes-no questions about object presence using the ``random'', ``popular'', and ``adversarial'' negative sampling methods. The accuracy of an MLLM on these 3,000 yes-no questions, denoted as the ``POPE acc'',  serves as a quantitative metric of its level of object hallucination. Therefore, the correlation between Pfram and POPE acc represents the degree to which Pfram can interpret object hallucinations of MLLMs\footnote{We also tested generative benchmarks like MMHal-Bench \cite{Sun2023AligningLM} and Object HalBench \cite{Rohrbach2018ObjectHI}. However, we found that these benchmarks strongly favor MLLMs that produce shorter responses, as noted in \cite{Zhai2023HallESwitchRA}. Since controlling response length for each MLLM trained on different data and formats is challenging, we opted to use only discriminative benchmarks in the paper.}.

\paragraph{Measuring Pframs of MLLMs.} 
To extract image representations in MLLMs, we use only the image as input. The input format varies depending on the specific MLLM, such as ``$\langle$s$\rangle$ $\langle$image\_token$\rangle$ $\times p$ $\langle$/s$\rangle$'', where $p$ represents the number of representation vectors for an image. We utilize the hidden states of the image (``$\langle$image\_token$\rangle$ $\times p$'') as image representations. 

For each model $\mathcal{M}$, we collect a set of layers $l$ whose output image representations are used to calculate Pfram as the representation system $\mathcal{F} = (\mathcal{M}, l)$. In our experiment, image representations are selected from every fourth layer of the LLM within the MLLM. For example, in a 40-layer LLM, we use representations from a set of layers $l \in S=\{0, 4, 8, ..., 40\}$ to calculate the Pfram scores separately, resulting in $11$ scores.

Based on these scores calculated from the selected layer representations of LLM, we consider three representation metrics to represent the overall Pfram score for an MLLM $\mathcal{M}$:
\begin{itemize}
    \item $\mathrm{Pfram_{input}}(\mathcal{M};\cdots):= \mathrm{Pfram}((\mathcal{M}, 0);\cdots)$, i.e., the Pfram score of layer 0's image representation;
    \item $\mathrm{Pfram_{mean}}(\mathcal{M};\cdots):=\frac{1}{|S|}\sum_{l\in S}\mathrm{Pfram}((\mathcal{M}, l);\cdots)$, i.e., the average Pfram score across all selected layers;
    \item $\mathrm{Pfram_{max}}(\mathcal{M};\cdots):=\max_{l\in S}\mathrm{Pfram}((\mathcal{M}, l);\cdots)$, i.e., the largest Pfram score from all selected layers.
\end{itemize}

Experiments are conducted using one NVIDIA A800 80G GPU, with each experiment taking approximately 6 hours.

\begin{figure}[t]
    \centering
    \includegraphics[width=0.48\textwidth]{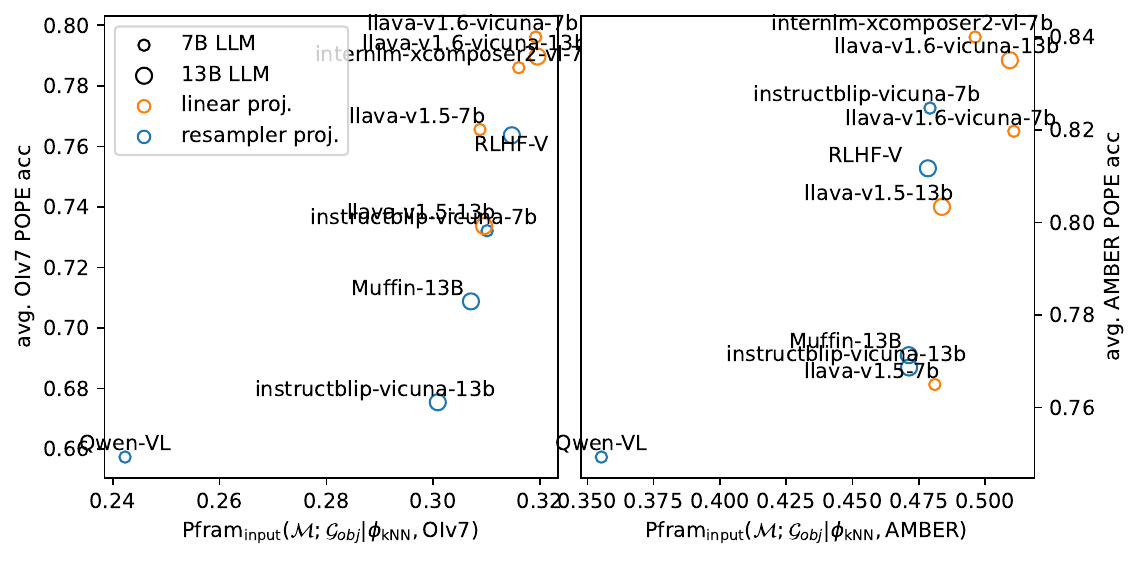}
    \includegraphics[width=0.48\textwidth]{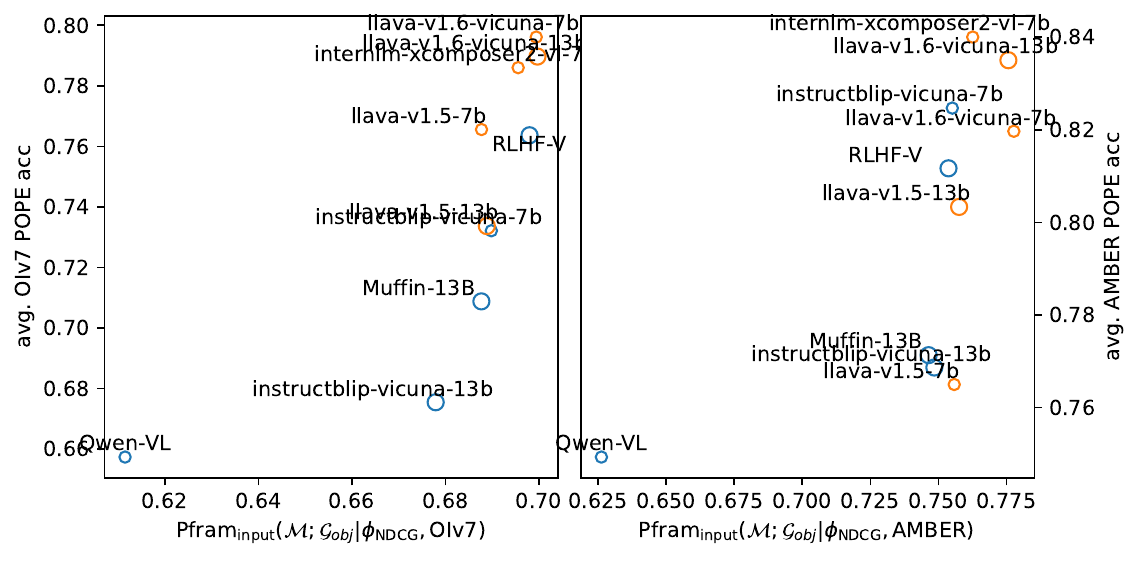}
    \includegraphics[width=0.48\textwidth]{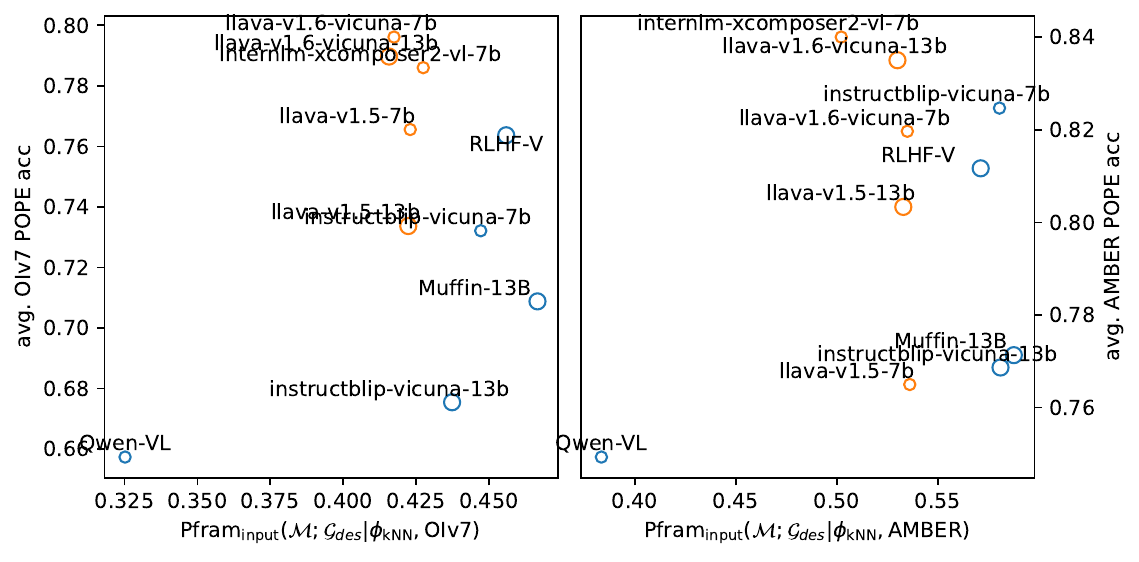}
    \includegraphics[width=0.48\textwidth]{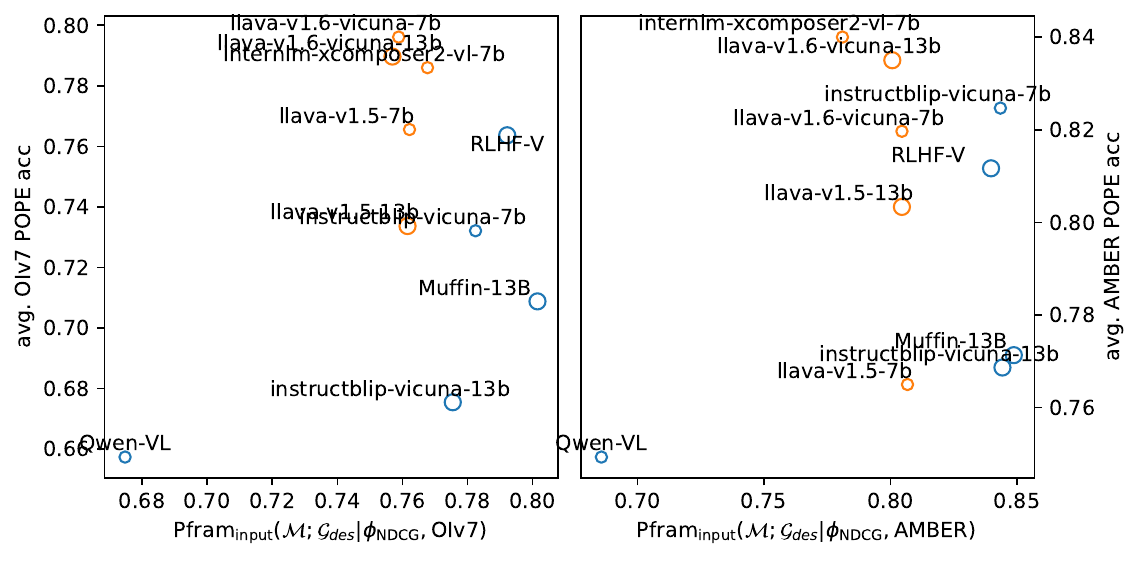}
    \caption{The POPE and Pfram scores for MLLMs with different LLM sizes (distinguished by point sizes) and different projectors (distinguished by point colors).}
    \label{fig:sor_scatter}
\end{figure}

\begin{table}[]
    \centering

    \begin{tabular}{c|c|cc}
    \toprule
     & Variable & $\mathcal{I}=$OIv7 & $\mathcal{I}=$AMBER \\
    \hline
    1 & $\mathrm{Pfram_{input}}(\mathcal{M};\mathcal{G}_{obj}| \phi_{\mathrm{kNN}}$) & \textbf{77.62 (0.01)} & \textbf{69.78 (0.02)} \\
    2 & $\mathrm{Pfram_{input}}(\mathcal{M};\mathcal{G}_{obj}| \phi_{\mathrm{NDCG}}$) & \textbf{77.15 (0.01)} &  \textbf{65.49 (0.04)} \\
    3 & $\mathrm{Pfram_{input}}(\mathcal{M};\mathcal{G}_{des}| \phi_{\mathrm{kNN}}$) & 35.14 (0.32) & 26.93 (0.45) \\
    4 & $\mathrm{Pfram_{input}}(\mathcal{M};\mathcal{G}_{des}| \phi_{\mathrm{NDCG}}$) & 36.52 (0.30) & 26.49 (0.46) \\
    5 & LLM Size & 14.37 (0.69) & 2.80 (0.94) \\
    6 & Projector Type & \textbf{72.78 (0.02)} & 44.44 (0.20) \\
    \bottomrule
    \end{tabular}
    \caption{The correlation coefficients between various variables and POPE acc on OIv7 and AMBER. To calculate the correlation coefficient and p value, for continuous variables (Pfram) we use Pearson's r, and for binary variables (LLM Size, Projector Type) we use Point Biserial \cite{Lev1949ThePB}. Statistically significant results with $p \le 0.05$ are highlighted in bold.}
    \label{tab:corr_pval}
\end{table}

\subsection{Pfram Has Strong Correlation with Object Hallucination}

In this section we empirically use $k=100$ for Mutual k-NN and NDCG@k and use $\mathrm{Pfram_{input}}(\mathcal{M})$ as an overall metric of a model $\mathcal{M}$. The results of POPE acc, $\mathrm{Pfram_{input}}(\mathcal{M}, \mathcal{G}_{obj})$ and $\mathrm{Pfram_{input}}(\mathcal{M}, \mathcal{G}_{des})$ are shown in Figure~\ref{fig:sor_scatter}, and the correlation coefficients between different variables and POPE acc are listed in Table \ref{tab:corr_pval}. Results of other $k$ and representation metrics are listed in the Appendix.

We have the following observations. Firstly, the Pfram($\mathcal{F}; \mathcal{G}_{obj}$) metric is statistically 
correlated with the POPE acc, as shown in rows 1-2 in Table \ref{tab:corr_pval}. This correlation is robust across different $\phi$ and $\mathcal{I}$. This demonstrates that Pfram($\mathcal{F}; \mathcal{G}_{obj}$) is an effective way to measure object information in image representations, and object information in image representations is indeed a crucial factor of object hallucinations of MLLMs.

Secondly, the hallucination level of MLLMs differs on different image datasets. For example, LLaVA-v1.5 7B performs the 4th best in 10 models in POPE acc with the OIv7 dataset, but performs the second worst (9th best in 10) in   POPE acc with the AMBER dataset. This suggests that we should be aware of the diversity of images and annotations when evaluating the hallucinations of MLLM, instead of simply using the images and annotations from a single source. This is an important issue that worth noting as most hallucination evaluation benchmarks \cite{Wang2023EvaluationAA,Li2023EvaluatingOH,Cha2024VisuallyDI,Rohrbach2018ObjectHI} only conduct experiments on one dataset, especially those that have already been widely used in training MLLMs like MSCOCO \cite{Lin2014MicrosoftCC} and Visual Genome \cite{Krishna2016VisualGC}. 

Thirdly, rows 3-6 show there is no strong and consistent correlation between POPE acc and other widely-studied characteristics of MLLMs, such as LLM size or the choice of modality projector. Notably, the similarity between image representations and short textual descriptions, which is quantified with Pfram($\mathcal{F}; \mathcal{G}_{des}$), does not correlate strongly with POPE acc.  This further demonstrates that improving the quality of image representations in terms of object information cannot be achieved by simply scaling model sizes, changing projectors or better aligning with short text descriptions, and it still remains to be an open question for future research.

Lastly, we observe from the second row of sub-figures in Figure \ref{fig:sor_scatter} that MLLMs with resampler projectors generally exhibit higher Pfram$(\mathcal{F}; \mathcal{G}_{des})$ values compared to those with linear projectors (i.e., the blue circles are positioned to the right of the orange ones, with the exception of Qwen-VL). This suggests that the image representations of resampler projectors align more closely with brief descriptions of images. We hypothesize that this phenomenon occurs because resampler projectors contain significantly more parameters than linear projectors, making them more susceptible to overfitting the training data, where image-description pairs constitute a substantial portion.

\section{Using Pfram to Diagnose MLLMs}

As previously demonstrated, Pfram$(\mathcal{F}; \mathcal{G}_{obj})$ shows a strong correlation with object hallucination in MLLMs.
Additionally, by inspecting different layers separately, we can use Pfram$(\mathcal{F}; \mathcal{G}_{obj})$ to diagnose MLLMs and explain the contribution of each component to object  hallucinations. \textbf{In this section, we specifically use Pfram$(\mathcal{F}; \mathcal{G}_{obj} | \phi_{\mathrm{NDCG}}, \mathrm{OIv7})$ and denote it as ``Pfram'' for simplicity.}

\subsection{Diagnosis of Different Modules in MLLMs} \label{sec:different_layers}

\begin{figure}
    \centering
    \includegraphics[width=0.6\textwidth]{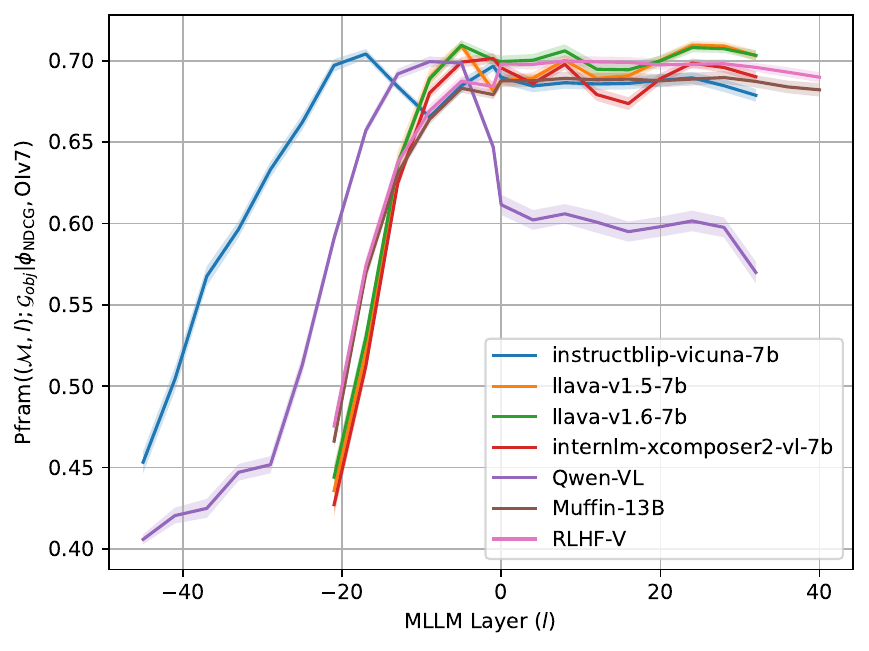}
    \caption{The change of Pfram with the layers in MLLM. X-axis denotes the layers in MLLM, where the left side (small number) is closer to input and the right side (large number) is closer to the output. Negative x-axis denotes the image hidden states from ViT and QFormer/Resampler (if any), x=0 denotes the image representations as the input of LLM, and positive x-asis denotes the image hidden states from the LLM. Y-axis is Pfram$(\mathcal{F}; \mathcal{G}_{obj} | \phi_{\mathrm{NDCG}}, \mathrm{OIv7})$, and the standard deviation is shown in shaded areas. Best view in color.}
    \label{fig:per_layer}
\end{figure}

Using the Pfram metric, we explore the image understanding process of MLLMs in greater detail. Specifically, we investigate how Pfram changes across different layers in the visual encoder and the LLM components of MLLMs. Do image representations gradually improve in their ability to discriminate objects across all layers? To address this, we measure the Pfram of different layers of MLLMs, as shown in Figure~\ref{fig:per_layer}. The results reveal distinct  differences between  modules in  function of extracting object information from images. Layers in visual encoders, including QFormers, consistently improve object recognition as the Pfram metric steadily increases. Conversely, LLMs do not primarily contribute to object recognition, as indicated by the plateau in the Pfram metric. We also observe a significant drop in Pfram from the -16th to the -12th layer in InstructBLIP (ViT to QFormer) and from the -8th to the 0th layer in Qwen-VL (ViT to LLM), suggesting potential issues with modality alignment that could contribute to object hallucinations.

By examining the differences in Pfram across various layers between Muffin-13B and RLHF-V, we can gain insights into the principles of hallucination-reduction training methods. RLHF-V is initialized from Muffin-13B, with all its parameters further refined using RLHF to encourage fine-grained segment-level correlations related to hallucinations. Compared to Muffin-13B, RLHF-V shows slight improvement in the higher layers of the visual encoder but exhibits significant improvement across all layers of the LLM in terms of Pfram. This demonstrates that the RLHF process effectively enhances the object-centered quality of image representations in LLMs, aligning them with fine-grained human preferences.

\subsection{Influence of Textual Instructions to Image Representations Depends on Model}

\begin{figure}
    \centering
    \includegraphics[width=0.7\textwidth]{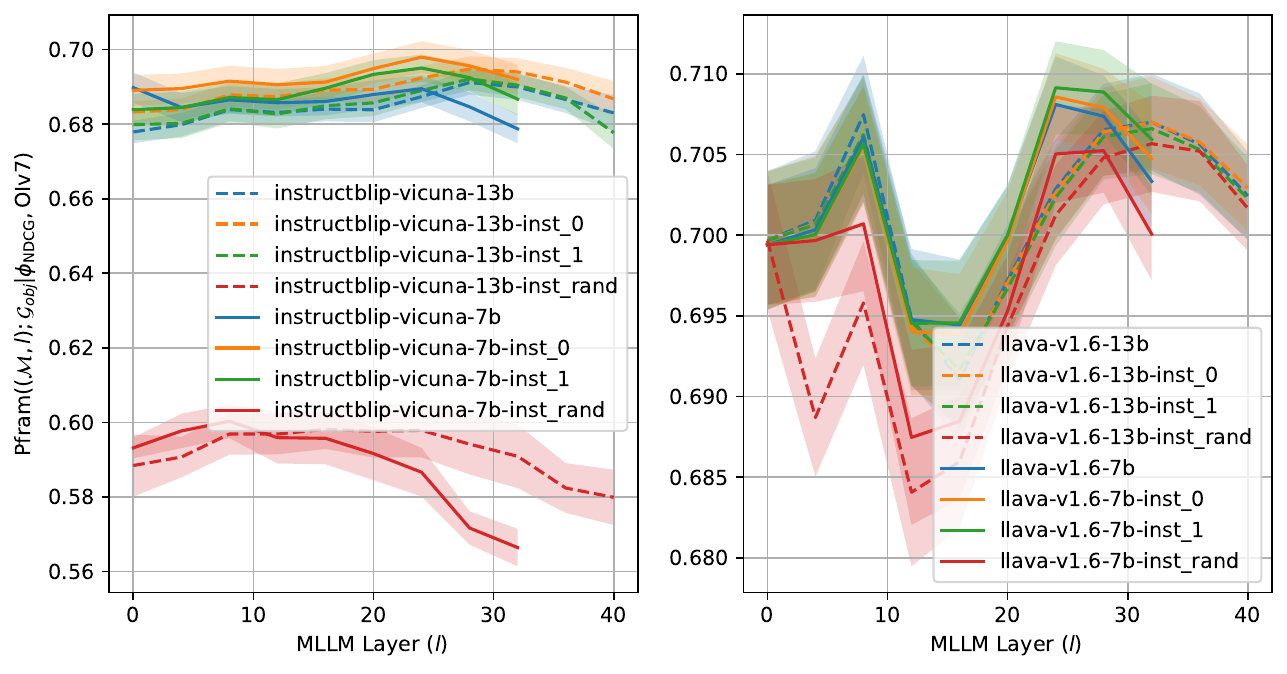}
    \caption{The change of Pfram with the layers in MLLM, when different textual instructions are conditioned on to acquire image representations. The definitions of axes are the same as Fig. \ref{fig:per_layer}. In the legend ``inst\_0'' and ``inst\_1'' denote the first and the second instruction is used, and ``inst\_rand'' denote the instruction is randomly chosen. Best view in color.}
    \label{fig:change_instruction}
\end{figure}

In previous sections, we explored how images are encoded in MLLMs when only the image is used as input. However, in real-world applications, an instruction is usually included as well. For example, in InstructBLIP, the instruction is used in the QFormer to help compress image information into query tokens, while in LLaVA, the instruction is prepended before the image as input to the LLM. In this section, we investigate two questions: 

\begin{itemize}
\item Do instructions affect the image representation? To explore this, we randomly choose either omitting instructions or  using a random instruction from a list for each image. We then verify whether image representations conditioned on different instructions can still retrieve images with common objects, assessing whether the Pfram metric changes.

\item Can instructions enhance object information in image representations, thereby reducing hallucinations in MLLMs? For this question, we calculate the Pfram metric using different instructions from a list across various experiments. In each experiment, the same instruction is used for encoding all images, allowing us to verify whether instructions can improve object-level information in image representations, as indicated by an increase in the Pfram metric. In these experiments, we use an ``instruction list'' containing two distinct instructions: one that promotes object-level image understanding, i.e., ``Identify the objects in this image'', and another commonly used for chat assistants, i.e., ``A chat between a curious user and an artificial intelligence assistant. The assistant gives helpful, detailed, and polite answers to the user's questions''.
\end{itemize}

Results are shown in Figure~\ref{fig:change_instruction}. We observe distinct patterns for InstructBLIP and LLaVA. In InstructBLIP, where instructions are used as QFormer input, we observe a significant drop in Pfram when different instructions are applied to different images (setting 1, red line). This indicates that different text instructions provide varied guidance to the image features in the representation space, making it challenging to directly calculate their similarities. When the same instruction is used for all images in a round of Pfram calculation (setting 2, orange and green lines), the Pfram metric improves slightly compared to when no instruction is used (blue line). These observations demonstrate that textual instructions in InstructBLIP's QFormer substantially impact object-level information processing of images.

In contrast, for LLaVA, where instructions are used as LLM input, the Pfram metric also decreases under setting 1 (red line), but the decline is not as significant as that in InstructBLIP. Additionally, under setting 2 (orange and green lines), the Pfram metric does not consistently improve. This suggests that in LLMs, the impact of textual instructions on image processing is  weaker than in QFormers.

\begin{figure}
    \centering
    \includegraphics[width=0.6\textwidth]{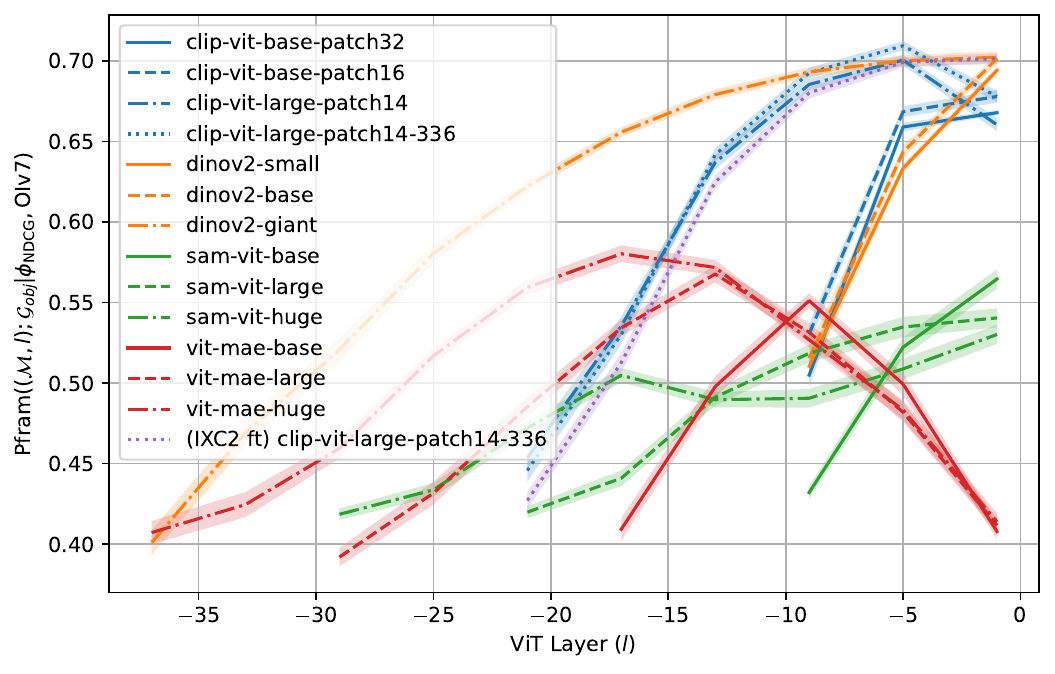}
    \caption{The change of Pfram with the layers in different ViTs. ``IXC2 ft'' denotes the ViT of the fine-tuned InternLM-XComposer2-VL model. The number of x-axis is the same as Fig. \ref{fig:per_layer} where -1 denotes the last layer of the ViT. Best view in color.}
    \label{fig:vits}
\end{figure}

\subsection{Jointly using Multiple ViTs May be Helpful}

Most open-source MLLMs employ various versions of CLIP-ViT \cite{Radford2021LearningTV,Sun2023EVACLIPIT} as their visual encoders, with only a few exceptions \cite{Alayrac2022FlamingoAV,Chen2023InternVLSU} so far. CLIP-ViT is popular because it is already aligned with text during its pre-training phase and can generate robust visual features for semantic similarity calculations. However, recent studies \cite{Tong2024EyesWS,Oquab2023DINOv2LR} suggest that CLIP has limitations in recognizing objects in complex images, which may lead to multimodal hallucination. To investigate whether alternative ViTs can mitigate these hallucinations, we calculate the Pfram metric for ViTs of different sizes and pre-training objectives: CLIP \cite{Radford2021LearningTV}, DINOv2 \cite{Oquab2023DINOv2LR}, SAM \cite{Kirillov2023SegmentA}, ViT-MAE \cite{He2021MaskedAA}, and a fine-tuned CLIP-ViT in IXC2 \cite{Dong2024InternLMXComposer2MF}. The results are presented in Figure~\ref{fig:vits}, leading to the following observations: 

\begin{itemize}
\item The pre-training objective has a more significant impact on Pfram scores than model sizes. For ViTs with the same pre-training objective, larger models generally achieve better Pfram scores, except for SAM.
\item For CLIP models of the same size, those with more image patches have higher Pfram scores.
\item  DINOv2 shows higher Pfram scores than CLIP in the last few layers, suggesting it might be a more effective ViT for reducing object hallucinations.

\end{itemize}
These findings indicate that different visual encoders excel in various aspects, and using multiple encoders together could be a promising approach for developing VLMs, as explored in \citet{Tong2024EyesWS,Lin2023SPHINXTJ,Gao2024SPHINXXSD}.

\section{Limitations}
The proposed Pfram metric has some limitations that warrant attention and require further investigation. The primary limitation is that Pfram is ordinal data rather than interval or ratio data, meaning it lacks a meaningful zero point and cannot be added or subtracted. This metric is only useful for comparing multiple models. Another significant limitation is that this work focuses solely on object-level discriminative hallucinations, while other types, such as attributes, relations, scenes, and generative hallucinations, are not yet included and should be considered as important future directions. 

\section{Conclusion}
In this paper, we explore the phenomenon of object hallucination in MLLMs. We introduce a parameter-free representation alignment method Pfram$(\mathcal{F}; \mathcal{G}| \phi, \mathcal{I})$, which evaluates a specific aspect of image representations by comparing their similarity with ground-truth labels. Our findings reveal that when using ground truth object annotations, Pfram$(\mathcal{F}; \mathcal{G}_{obj})$ exhibits a strong  correlation with the level of object hallucination across various state-of-the-art MLLMs, regardless of model structures and model sizes. Additionally, we investigate the contribution of each component in the image understanding process in MLLMs with Pfram scores and discuss limitations and future research directions.

\bibliography{iclr2025_conference}
\bibliographystyle{iclr2025_conference}

\appendix

\section{MLLMs used In the Experiments}
Table \ref{tab:models} lists the MLLMs used in the experiments in Section 3 in the main paper. These MLLMs use a wide range of ViTs \cite{Radford2021LearningTV,Fang2022EVAET,OpenCLIP2021,beit3} and LLMs \cite{Vicuna2024Lmsys,Cai2024InternLM2TR,Bai2023QwenTR}.

\section{Formal Descriptions of Mutual k-NN and NDCG@k}
The formal descriptions of Mutual k-NN and NDCG@k are shown in Algorithm \ref{alg:ndcg} and Algorithm \ref{alg:knn}.

\renewcommand{\algorithmicrequire}{\textbf{Input:}}
\renewcommand{\algorithmicensure}{\textbf{Output:}}
\begin{algorithm}[t]
\caption{Calculate NDCG@k for each anchor image}
\label{alg:ndcg}
\begin{algorithmic}
\Require 
  \State List of relevance scores for each target image calculated by $\mathcal{G}$: $\mathbf{rel^\mathcal{G}} = [rel^\mathcal{G}, rel^\mathcal{G}_2, \ldots, rel^\mathcal{G}_n]$
  \State List of target image indices ranked by the system to evaluate $\mathcal{F}$: $\mathbf{rank^\mathcal{F}} = [rank^\mathcal{F}_1, rank^\mathcal{F}_2, \ldots, rank^\mathcal{F}_n]$
  \State The number of the first few objects to calculate $\mathbf{k} \le n$
\Ensure NDCG@k
\smallskip
\State According to $\mathbf{rel}$, get an ideal ranking $\mathbf{rank^\mathcal{G}} = [rank^\mathcal{G}_1, rank^\mathcal{G}_2, \ldots, rank^\mathcal{G}_n] (rel_{rank^\mathcal{G}_1} \ge rel_{rank^\mathcal{G}_2} \ge \ldots \ge rel_{rank^\mathcal{G}_n})$
\State $DCG@k = \sum_{j=1}^k \frac{rel_{rank^\mathcal{F}_j}}{\log_2(j+1)}$
\State $IDCG@k = \sum_{i=j}^k \frac{rel_{rank^\mathcal{G}_j}}{\log_2(j+1)}$
\State $NDCG@k = \frac{DCG@k}{IDCG@k}$
\end{algorithmic}
\end{algorithm}

\begin{algorithm}[t]
\caption{Calculate Mutual k-NN for each anchor image}
\label{alg:knn}
\begin{algorithmic}
\Require 
  \State List of target image indices ranked by the ground truth representation system $\mathcal{G}$: $\mathbf{rank^\mathcal{G}} = [rank^\mathcal{G}_1, rank^\mathcal{G}_2, \ldots, rank^\mathcal{G}_n]$
  \State List of target image indices ranked by the system to evaluate $\mathcal{F}$: $\mathbf{rank^\mathcal{F}} = [rank^\mathcal{F}_1, rank^\mathcal{F}_2, \ldots, rank^\mathcal{F}_n]$
  \State The number of the first few objects to calculate $\mathbf{k} \le n$
\Ensure Mutual k-NN
\smallskip
\State Get the first $k$ items in each list:
\State $\mathbf{rank^\mathcal{G}@k} = [rank^\mathcal{G}_1, rank^\mathcal{G}_2, \ldots, rank^\mathcal{G}_k]$
\State $\mathbf{rank^\mathcal{F}@k} = [rank^\mathcal{F}_1, rank^\mathcal{F}_2, \ldots, rank^\mathcal{F}_k]$
\State Mutual k-NN $= \frac{|\mathbf{rank^\mathcal{G}@k} \cap \mathbf{rank^\mathcal{F}@k} |}{k}$
\end{algorithmic}
\end{algorithm}

\begin{table}[t]
    \centering
    \resizebox{\textwidth}{!}{
    \begin{tabular}{c|ccc|c}
    \toprule
    \multirow{2}{*}{Model} & \multicolumn{3}{c|}{Structure} & \multirow{2}{*}{Training} \\
     & ViT & Projector & LLM & \\
    \hline
    instructblip-vicuna-7b \cite{Dai2023InstructBLIPTG} & EVA-CLIP ViT-g/14 \snow & 12-Layer Resampler \fire & Vicuna 7B v1.1 \snow & P+M \\
    instructblip-vicuna-13b \cite{Dai2023InstructBLIPTG} & EVA-CLIP ViT-g/14 \snow & 12-Layer Resampler \fire & Vicuna 13B v1.1 \snow & P+M \\
    llava-v1.5-7b \cite{Liu2023ImprovedBW} & CLIP ViT-L/14@336p \snow  & 2-Layer Linear \fire & Vicuna 7B v1.5 \fire & P+M \\
    llava-v1.5-13b \cite{Liu2023ImprovedBW} & CLIP ViT-L/14@336p \snow & 2-Layer Linear \fire & Vicuna 13B v1.5 \fire & P+M \\
    internlm-xcomposer2-vl-7b \cite{Dong2024InternLMXComposer2MF} & CLIP ViT-L/14@336p \fire  & 2-Layer Linear \fire & InternLM2-7B \fire & P+M \\
    Muffin-13B \cite{Yu2023ReformulatingVF} & \multicolumn{2}{c}{BEiT3-L/16@672p \fire} & Vicuna 13B v1.5 \fire & P+M \\
    RLHF-V (13B) \cite{Yu2023RLHFVTT} & \multicolumn{2}{c}{BEiT3-L/16@672p \fire} & Vicuna 13B v1.5 \fire & P+M+R \\
    llava-v1.6-vicuna-7b \cite{liu2024llava} & CLIP ViT-L/14@336p \snow & 2-Layer Linear \fire & Vicuna 7B v1.5 \fire & P+M \\
    llava-v1.6-vicuna-13b \cite{liu2024llava} & CLIP ViT-L/14@336p \snow & 2-Layer Linear \fire & Vicuna 7B v1.5 \fire & P+M \\
    Qwen-VL (7B) \cite{Bai2023QwenVLAF} & Openclip ViT-bigG \fire & 1-Layer Resampler \fire & Qwen-LM 7B \fire & P+M \\
    \bottomrule
    \end{tabular}}
    \caption{MLLMs used in the experiments. In the ``Structure'' columns, snow denotes all parameters in this module are frozen throughout the entire training process, and fire denotes some parameters in this module are trained in some of the training steps. In the ``Training'' column, ``P'' denotes a pre-training phase, ``M'' denotes a multi-task training phase, and ``R'' denotes a RLHF training phase.}
    \label{tab:models}
\end{table}

\section{Difference Between Using Object Annotations and Brief Descriptions as Ground Truth}
\begin{figure}
    \centering
    \includegraphics[width=\textwidth]{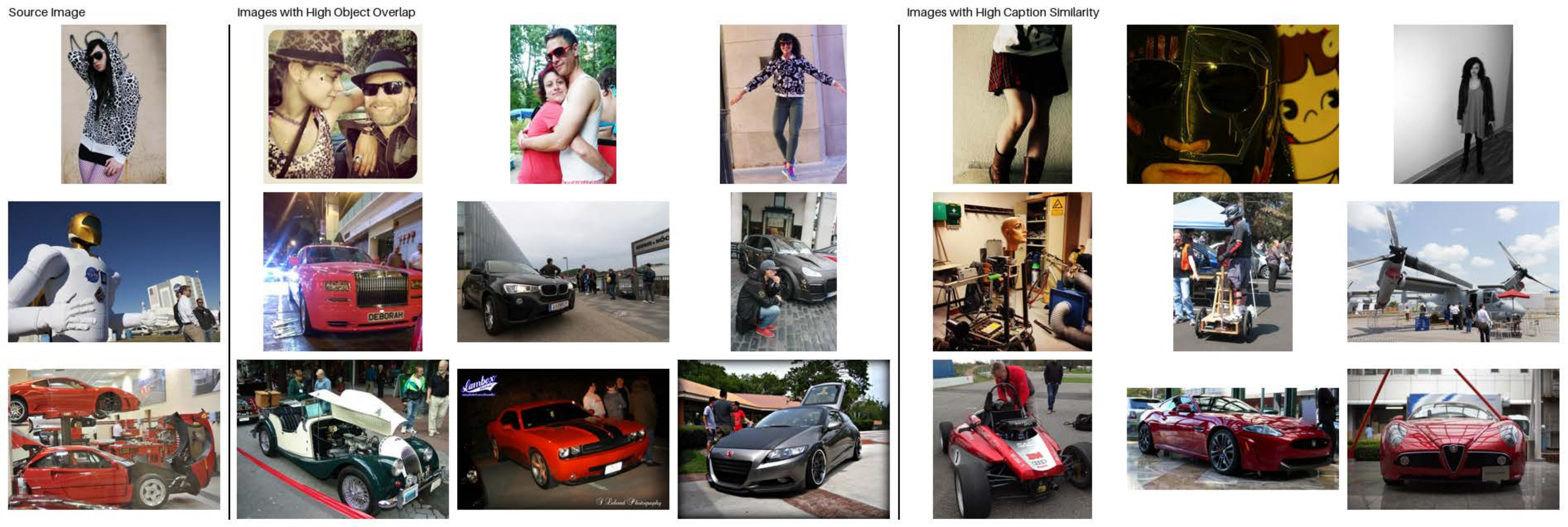}
    \caption{Examples of the nearest neighbouring images retrieved using object labels ($\mathcal{G}_{obj}$, left) and captions ($\mathcal{G}_{des}$, right).}
    \label{fig:retrieved_images}
\end{figure}

Fig. \ref{fig:retrieved_images} shows the difference of the nearest neighboring images retrieved using object labels ($\mathcal{G}_{obj}$) and brief descriptions ($\mathcal{G}_{des}$). Though they look roughly similar in theme or style, images retrieved with object labels have more similar details with the source images, e.g., women and sunglasses in the first example, men and buildings in the background in the second example, and visible wheels and tires in the third example.

\section{Detailed Results of Pfram and POPE}
The detailed results of Pfram and its correlation coefficient with POPE acc are listed in Table \ref{tab:OIv7_SOR}-\ref{tab:AMBER_SC_kNN}.

\begin{table}[t]
    \centering
    \resizebox{\textwidth}{!}{
    \begin{tabular}{c|c|c|c|c|c|c|c}
    \toprule
    \multirow{2}{*}{Model} & OIv7 POPE & \multicolumn{6}{c}{Pfram($\mathcal{F}; \mathcal{G}_{obj}| \phi_{\mathrm{NDCG}}, $ OIv7) (std.)} \\
     & acc (std.) & k=25 input & k=25 mean & k=25 max & k=100 input & k=100 mean & k=100 max \\
     \hline
     instructblip-vicuna-7b & 73.21 (1.29) & 67.84 (0.45) & 67.54 (0.44) & 68.02 (0.45) & 68.98 (0.41) & 68.59 (0.38) & 69.01 (0.37) \\
    instructblip-vicuna-13b & 67.54 (1.41) & 66.93 (0.45) & 67.41 (0.49) & 68.02 (0.50) & 67.79 (0.30) & 68.46 (0.33) & 69.12 (0.33) \\
    llava-v1.5-7b & 76.56 (1.14) & 67.60 (0.42) & 68.41 (0.43) & 69.39 (0.38) & 68.77 (0.34) & 69.77 (0.33) & 70.94 (0.27) \\
    llava-v1.5-13b & 73.37 (1.34) & 67.73 (0.49) & 68.41 (0.43) & 69.23 (0.32) & 68.88 (0.40) & 69.82 (0.36) & 70.81 (0.28) \\
    internlm-xcomposer2-vl-7b & 78.60 (1.46) & 68.57 (0.40) & 67.74 (0.36) & 68.62 (0.35) & 69.55 (0.34) & 68.95 (0.33) & 69.87 (0.31) \\
    Muffin-13B & 70.88 (1.60) & 67.89 (0.46) & 67.89 (0.49) & 68.15 (0.54) & 68.77 (0.39) & 68.73 (0.39) & 68.97 (0.39) \\
    RLHF-V & 76.37 (1.48) & 68.97 (0.37) & 68.94 (0.42) & 69.15 (0.43) & 69.80 (0.31) & 69.69 (0.30) & 69.98 (0.31) \\
    llava-v1.6-vicuna-7b & 79.61 (1.85) & 68.66 (0.44) & 68.85 (0.44) & 69.58 (0.39) & 69.94 (0.37) & 70.15 (0.35) & 70.81 (0.30) \\
    llava-v1.6-vicuna-13b & 78.97 (1.44) & 68.72 (0.52) & 68.77 (0.45) & 69.32 (0.44) & 69.97 (0.43) & 70.14 (0.34) & 70.79 (0.36) \\
    Qwen-VL & 65.73 (0.51) & 61.19 (0.59) & 60.16 (0.55) & 61.19 (0.59) & 61.15 (0.62) & 59.80 (0.62) & 61.15 (0.62) \\
    \hline
    Pearson's r (p value) & \textcolor{gray}{100.00 (0.00)} & 76.37 (0.01)* & 70.76 (0.02)* & 73.36 (0.02)* & 77.15 (0.01)* & 71.52 (0.02)* & 73.48 (0.02)* \\
    \bottomrule
    \toprule
    \multirow{2}{*}{Model} & OIv7 POPE & \multicolumn{6}{c}{Pfram($\mathcal{F}; \mathcal{G}_{obj}| \phi_{\mathrm{NDCG}}, $ OIv7) (std.)} \\
     & acc (std.) & k=400 input & k=400 mean & k=400 max & k=1600 input & k=1600 mean & k=1600 max \\
    \hline
    instructblip-vicuna-7b & 73.21 (1.29) & 73.95 (0.56) & 73.23 (0.41) & 73.95 (0.56) & 91.28 (0.14) & 91.06 (0.10) & 91.28 (0.14) \\
    instructblip-vicuna-13b & 67.54 (1.41) & 72.64 (0.38) & 73.16 (0.38) & 73.60 (0.29) & 90.90 (0.10) & 91.03 (0.09) & 91.20 (0.08) \\
    llava-v1.5-7b & 76.56 (1.14) & 73.38 (0.32) & 74.84 (0.35) & 76.25 (0.31) & 91.11 (0.07) & 91.54 (0.07) & 91.98 (0.06) \\
    llava-v1.5-13b & 73.37 (1.34) & 73.36 (0.39) & 74.85 (0.36) & 76.12 (0.29) & 91.08 (0.09) & 91.52 (0.08) & 91.92 (0.05) \\
    internlm-xcomposer2-vl-7b & 78.60 (1.46) & 73.85 (0.35) & 73.66 (0.38) & 75.01 (0.33) & 91.18 (0.09) & 91.10 (0.09) & 91.58 (0.08) \\
    Muffin-13B & 70.88 (1.60) & 73.09 (0.49) & 72.94 (0.49) & 73.30 (0.49) & 91.08 (0.11) & 91.06 (0.11) & 91.15 (0.11) \\
    RLHF-V & 76.37 (1.48) & 74.11 (0.35) & 73.73 (0.35) & 74.35 (0.35) & 91.40 (0.08) & 91.32 (0.08) & 91.48 (0.08) \\
    llava-v1.6-vicuna-7b & 79.61 (1.85) & 74.93 (0.35) & 75.20 (0.35) & 75.92 (0.37) & 91.52 (0.08) & 91.63 (0.08) & 91.86 (0.07) \\
    llava-v1.6-vicuna-13b & 78.97 (1.44) & 74.85 (0.39) & 75.20 (0.35) & 76.05 (0.34) & 91.48 (0.09) & 91.61 (0.08) & 91.86 (0.09) \\
    Qwen-VL & 65.73 (0.51) & 65.23 (0.74) & 63.72 (0.78) & 65.23 (0.74) & 88.31 (0.20) & 87.85 (0.20) & 88.31 (0.20) \\
    \hline
    Pearson's r (p value) & \textcolor{gray}{100.00 (0.00)} & 76.17 (0.01)* & 72.65 (0.02)* & 76.10 (0.01)* & 72.62 (0.02)* & 70.53 (0.02)* & 74.07 (0.01)* \\
    \bottomrule
    \end{tabular}}
    \caption{Results of OIv7 POPE acc and Pfram($\mathcal{F}; \mathcal{G}_{obj}| \phi_{\mathrm{NDCG}}, $ OIv7) scores with different $k$ and represenatation metrics. Mean values and standard deviation of 3 experiments are reported. In the last row we report the Pearson's r correlation (and p value) between each column and POPE acc. *: Statistically significant with $p \le 0.05$.} \label{tab:OIv7_SOR}
\end{table}

\begin{table}[t]
    \centering
    \resizebox{\textwidth}{!}{
    \begin{tabular}{c|c|c|c|c|c|c|c}

    \toprule
    \multirow{2}{*}{Model} & OIv7 POPE & \multicolumn{6}{c}{Pfram($\mathcal{F}; \mathcal{G}_{des}| \phi_{\mathrm{NDCG}}, $ OIv7) (std.)} \\
     & acc (std.) & k=25 input & k=25 mean & k=25 max & k=100 input & k=100 mean & k=100 max \\
    \hline
    instructblip-vicuna-7b & 73.21 (1.29) & 78.94 (0.29) & 79.29 (0.35) & 79.77 (0.40) & 78.25 (0.40) & 78.71 (0.39) & 79.09 (0.39) \\
    instructblip-vicuna-13b & 67.54 (1.41) & 78.13 (0.24) & 78.71 (0.36) & 79.20 (0.42) & 77.55 (0.33) & 78.13 (0.41) & 78.48 (0.46) \\
    llava-v1.5-7b & 76.56 (1.14) & 76.63 (0.27) & 76.30 (0.25) & 77.31 (0.23) & 76.22 (0.34) & 75.88 (0.32) & 76.96 (0.32) \\
    llava-v1.5-13b & 73.37 (1.34) & 76.60 (0.25) & 76.11 (0.24) & 77.22 (0.21) & 76.16 (0.35) & 75.71 (0.32) & 76.93 (0.32) \\
    internlm-xcomposer2-vl-7b & 78.60 (1.46) & 76.99 (0.20) & 77.42 (0.13) & 79.32 (0.10) & 76.77 (0.32) & 77.21 (0.26) & 78.94 (0.26) \\
    Muffin-13B & 70.88 (1.60) & 80.66 (0.21) & 80.88 (0.20) & 81.12 (0.18) & 80.15 (0.33) & 80.32 (0.33) & 80.52 (0.34) \\
    RLHF-V & 76.37 (1.48) & 80.00 (0.31) & 80.30 (0.27) & 80.53 (0.22) & 79.22 (0.35) & 79.38 (0.37) & 79.84 (0.32) \\
    llava-v1.6-vicuna-7b & 79.61 (1.85) & 76.14 (0.20) & 76.81 (0.17) & 77.72 (0.15) & 75.88 (0.27) & 76.47 (0.26) & 77.10 (0.27) \\
    llava-v1.6-vicuna-13b & 78.97 (1.44) & 76.04 (0.18) & 76.52 (0.18) & 77.11 (0.18) & 75.69 (0.27) & 76.12 (0.28) & 76.54 (0.29) \\
    Qwen-VL & 65.73 (0.51) & 69.12 (0.69) & 68.49 (0.65) & 69.77 (0.60) & 67.48 (0.87) & 66.47 (0.82) & 67.48 (0.87) \\
    \hline
    Pearson's r (p value) & \textcolor{gray}{100.00 (0.00)} & 30.11 (0.40) & 33.42 (0.35) & 38.85 (0.27) & 36.52 (0.30) & 39.36 (0.26) & 43.99 (0.20) \\
    \bottomrule
    \toprule
    \multirow{2}{*}{Model} & OIv7 POPE & \multicolumn{6}{c}{Pfram($\mathcal{F}; \mathcal{G}_{des}| \phi_{\mathrm{NDCG}}, $ OIv7) (std.)} \\
     & acc (std.) & k=400 input & k=400 mean & k=400 max & k=1600 input & k=1600 mean & k=1600 max \\
    \hline
    instructblip-vicuna-7b & 73.21 (1.29) & 78.26 (0.43) & 78.69 (0.35) & 79.05 (0.32) & 93.51 (0.12) & 93.60 (0.11) & 93.69 (0.11) \\
    instructblip-vicuna-13b & 67.54 (1.41) & 77.95 (0.35) & 78.12 (0.35) & 78.43 (0.32) & 93.42 (0.11) & 93.46 (0.12) & 93.54 (0.11) \\
    llava-v1.5-7b & 76.56 (1.14) & 77.00 (0.37) & 76.90 (0.35) & 77.77 (0.33) & 93.15 (0.13) & 93.09 (0.11) & 93.34 (0.11) \\
    llava-v1.5-13b & 73.37 (1.34) & 77.05 (0.36) & 76.82 (0.34) & 77.84 (0.33) & 93.16 (0.12) & 93.06 (0.11) & 93.35 (0.11) \\
    internlm-xcomposer2-vl-7b & 78.60 (1.46) & 77.73 (0.34) & 77.86 (0.30) & 79.37 (0.32) & 93.28 (0.12) & 93.34 (0.10) & 93.80 (0.10) \\
    Muffin-13B & 70.88 (1.60) & 80.06 (0.40) & 80.00 (0.40) & 80.26 (0.38) & 94.04 (0.12) & 94.03 (0.12) & 94.09 (0.11) \\
    RLHF-V & 76.37 (1.48) & 79.20 (0.41) & 78.86 (0.44) & 79.72 (0.39) & 93.84 (0.12) & 93.77 (0.13) & 93.97 (0.12) \\
    llava-v1.6-vicuna-7b & 79.61 (1.85) & 77.25 (0.32) & 77.50 (0.30) & 77.78 (0.27) & 93.15 (0.11) & 93.25 (0.11) & 93.38 (0.10) \\
    llava-v1.6-vicuna-13b & 78.97 (1.44) & 77.02 (0.31) & 77.16 (0.30) & 77.40 (0.29) & 93.10 (0.11) & 93.15 (0.11) & 93.25 (0.10) \\
    Qwen-VL & 65.73 (0.51) & 68.05 (0.87) & 67.16 (0.83) & 68.05 (0.87) & 90.69 (0.25) & 90.45 (0.23) & 90.71 (0.23) \\
    \hline
    Pearson's r (p value) & \textcolor{gray}{100.00 (0.00)} & 47.11 (0.17) & 48.58 (0.15) & 51.33 (0.13) & 44.03 (0.20) & 46.63 (0.17) & 51.00 (0.13) \\
    \bottomrule
    \end{tabular}}
    \caption{Results of OIv7 POPE acc and Pfram($\mathcal{F}; \mathcal{G}_{des}| \phi_{\mathrm{NDCG}}, $ OIv7) scores. Mean values and standard deviation of 3 experiments are reported. In the last row we report the Pearson's r correlation (and p value) between each column and OIv7 POPE acc.} \label{tab:OIv7_SC}
\end{table}

\begin{table}[t]
    \centering
    \resizebox{\textwidth}{!}{
    \begin{tabular}{c|c|c|c|c|c|c|c}
    \toprule
    \multirow{2}{*}{Model} & AMBER POPE & \multicolumn{6}{c}{Pfram($\mathcal{F}; \mathcal{G}_{obj}| \phi_{\mathrm{NDCG}},$ AMBER)} \\
     & acc & k=25 input & k=25 mean & k=25 max & k=100 input & k=100 mean & k=100 max \\
    \hline
    instructblip-vicuna-7b & 82.47 & 76.70 & 77.11 & 77.76 & 75.51 & 76.09 & 76.92 \\
    instructblip-vicuna-13b & 76.87 & 76.15 & 76.33 & 76.98 & 74.85 & 75.21 & 75.95 \\
    llava-v1.5-7b & 76.50 & 76.17 & 77.68 & 80.13 & 75.58 & 78.00 & 81.34 \\
    llava-v1.5-13b & 80.33 & 76.12 & 77.90 & 80.09 & 75.76 & 78.71 & 81.66 \\
    internlm-xcomposer2-vl-7b & 84.00 & 76.41 & 76.50 & 78.32 & 76.26 & 76.67 & 79.33 \\
    Muffin-13B & 77.13 & 76.07 & 76.06 & 76.46 & 74.64 & 74.49 & 74.87 \\
    RLHF-V & 81.17 & 77.37 & 77.15 & 77.43 & 75.37 & 74.83 & 75.40 \\
    llava-v1.6-vicuna-7b & 81.97 & 77.28 & 78.06 & 79.81 & 77.77 & 78.84 & 80.82 \\
    llava-v1.6-vicuna-13b & 83.50 & 76.81 & 78.25 & 79.86 & 77.57 & 79.34 & 81.15 \\
    Qwen-VL & 74.93 & 66.70 & 66.90 & 69.45 & 62.62 & 62.28 & 64.03 \\
    \hline
    Pearson's r (p value) & \textcolor{gray}{100.00 (0.00)} & 60.27 (0.07) & 58.93 (0.07) & 57.68 (0.08) & 65.49 (0.04)* & 60.94 (0.06) & 58.39 (0.08) \\
    \bottomrule

    \toprule
    \multirow{2}{*}{Model} & AMBER POPE & \multicolumn{6}{c}{Pfram($\mathcal{F}; \mathcal{G}_{obj}| \phi_{\mathrm{NDCG}},$ AMBER)} \\
    & acc & k=400 input & k=400 mean & k=400 max & k=639 input & k=639 mean & k=639 max \\
    \hline
    instructblip-vicuna-7b & 82.47 & 84.53 & 84.91 & 85.59 & 91.56 & 91.74 & 92.03 \\
    instructblip-vicuna-13b & 76.87 & 84.43 & 84.60 & 85.09 & 91.41 & 91.50 & 91.76 \\
    llava-v1.5-7b & 76.50 & 85.24 & 87.15 & 89.47 & 91.58 & 92.38 & 93.53 \\
    llava-v1.5-13b & 80.33 & 85.24 & 87.59 & 89.85 & 91.60 & 92.57 & 93.61 \\
    internlm-xcomposer2-vl-7b & 84.00 & 83.89 & 84.72 & 87.72 & 91.50 & 91.67 & 92.68 \\
    Muffin-13B & 77.13 & 83.57 & 83.51 & 83.81 & 91.22 & 91.18 & 91.32 \\
    RLHF-V & 81.17 & 83.91 & 83.33 & 84.06 & 91.51 & 91.32 & 91.54 \\
    llava-v1.6-vicuna-7b & 81.97 & 86.82 & 87.58 & 88.93 & 92.24 & 92.57 & 93.29 \\
    llava-v1.6-vicuna-13b & 83.50 & 86.57 & 87.94 & 89.28 & 92.13 & 92.72 & 93.38 \\
    Qwen-VL & 74.93 & 72.02 & 71.64 & 72.48 & 86.57 & 86.44 & 87.07 \\
    \hline
    Pearson's r (p value) & \textcolor{gray}{100.00 (0.00)} & 57.82 (0.08) & 55.58 (0.10) & 57.58 (0.08) & 61.05 (0.06) & 57.95 (0.08) & 57.06 (0.08) \\
    \bottomrule
    \end{tabular}}
    \caption{Results of AMBER POPE acc and Pfram($\mathcal{F}; \mathcal{G}_{obj}| \phi_{\mathrm{NDCG}},$ AMBER) scores. In the last row we report the Pearson's r correlation (and p value) between each column and AMBER POPE acc. *: Statistically significant with $p \le 0.05$.} \label{tab:AMBER_SOR}
\end{table}

\begin{table}[t]
    \centering
    \resizebox{\textwidth}{!}{
    \begin{tabular}{c|c|c|c|c|c|c|c}
    \toprule
    \multirow{2}{*}{Model} & AMBER POPE & \multicolumn{6}{c}{Pfram($\mathcal{F}; \mathcal{G}_{des}| \phi_{\mathrm{NDCG}},$ AMBER)} \\
     & acc & k=25 input & k=25 mean & k=25 max & k=100 input & k=100 mean & k=100 max \\
    \hline
    instructblip-vicuna-7b & 82.47 & 84.93 & 85.17 & 85.63 & 84.34 & 84.57 & 85.00 \\
    instructblip-vicuna-13b & 76.87 & 84.85 & 84.83 & 85.14 & 84.42 & 84.34 & 84.52 \\
    llava-v1.5-7b & 76.50 & 80.47 & 79.96 & 81.36 & 80.67 & 80.28 & 81.51 \\
    llava-v1.5-13b & 80.33 & 80.22 & 79.21 & 80.99 & 80.45 & 79.70 & 81.34 \\
    internlm-xcomposer2-vl-7b & 84.00 & 78.03 & 78.87 & 81.26 & 78.10 & 79.05 & 81.67 \\
    Muffin-13B & 77.13 & 85.33 & 85.50 & 85.76 & 84.87 & 84.94 & 85.17 \\
    RLHF-V & 81.17 & 84.96 & 85.09 & 85.26 & 83.97 & 83.65 & 84.28 \\
    llava-v1.6-vicuna-7b & 81.97 & 79.85 & 80.01 & 80.82 & 80.45 & 80.38 & 80.97 \\
    llava-v1.6-vicuna-13b & 83.50 & 79.41 & 79.54 & 80.21 & 80.07 & 79.98 & 80.49 \\
    Qwen-VL & 74.93 & 71.15 & 71.40 & 73.64 & 68.58 & 68.17 & 69.47 \\
    \hline
    Pearson's r (p value) & \textcolor{gray}{100.00 (0.00)} & 16.69 (0.64) & 20.43 (0.57) & 22.86 (0.53) & 26.49 (0.46) & 30.72 (0.39) & 36.39 (0.30) \\
    \bottomrule

    \toprule
    \multirow{2}{*}{Model} & AMBER POPE & \multicolumn{6}{c}{Pfram($\mathcal{F}; \mathcal{G}_{des}| \phi_{\mathrm{NDCG}},$ AMBER)} \\
    & acc & k=400 input & k=400 mean & k=400 max & k=639 input & k=639 mean & k=639 max \\
    \hline
    instructblip-vicuna-7b & 82.47 & 86.88 & 86.99 & 87.39 & 94.85 & 94.92 & 95.05 \\
    instructblip-vicuna-13b & 76.87 & 87.28 & 87.09 & 87.32 & 94.93 & 94.88 & 94.93 \\
    llava-v1.5-7b & 76.50 & 85.62 & 85.62 & 86.25 & 93.84 & 93.76 & 94.12 \\
    llava-v1.5-13b & 80.33 & 85.42 & 85.28 & 86.06 & 93.78 & 93.58 & 94.04 \\
    internlm-xcomposer2-vl-7b & 84.00 & 83.22 & 83.80 & 85.73 & 93.00 & 93.26 & 94.05 \\
    Muffin-13B & 77.13 & 87.52 & 87.43 & 87.58 & 95.06 & 95.06 & 95.12 \\
    RLHF-V & 81.17 & 86.81 & 86.22 & 87.06 & 94.82 & 94.70 & 94.92 \\
    llava-v1.6-vicuna-7b & 81.97 & 85.74 & 85.64 & 86.01 & 93.78 & 93.77 & 93.96 \\
    llava-v1.6-vicuna-13b & 83.50 & 85.40 & 85.38 & 85.69 & 93.65 & 93.64 & 93.79 \\
    Qwen-VL & 74.93 & 76.04 & 75.76 & 76.32 & 90.20 & 90.10 & 90.50 \\
    \hline
    Pearson's r (p value) & \textcolor{gray}{100.00 (0.00)} & 31.76 (0.37) & 35.49 (0.31) & 42.71 (0.22) & 25.98 (0.47) & 30.22 (0.40) & 36.22 (0.30) \\
    \bottomrule
    \end{tabular}}
    \caption{Results of AMBER POPE acc and Pfram($\mathcal{F}; \mathcal{G}_{des}| \phi_{\mathrm{NDCG}},$ AMBER) scores. In the last row we report the Pearson's r correlation (and p value) between each column and AMBER POPE acc.} \label{tab:AMBER_SC}
\end{table}

\begin{table}[t]
    \centering
    \resizebox{\textwidth}{!}{
    \begin{tabular}{c|c|c|c|c|c|c|c}
    \toprule
    \multirow{2}{*}{Model} & OIv7 POPE & \multicolumn{6}{c}{Pfram($\mathcal{F}; \mathcal{G}_{obj}| \phi_{\mathrm{kNN}},$ OIv7) (std.)} \\
    & acc (std.) & k=25 input & k=25 mean & k=25 max & k=100 input & k=100 mean & k=100 max \\
    \hline
    instructblip-vicuna-7b & 73.21 (1.29) & 20.66 (0.28) & 20.57 (0.25) & 20.99 (0.26) & 31.01 (0.16) & 30.77 (0.06) & 31.10 (0.10) \\
    instructblip-vicuna-13b & 67.54 (1.41) & 20.01 (0.30) & 20.51 (0.40) & 21.04 (0.45) & 30.09 (0.07) & 30.67 (0.09) & 31.15 (0.13) \\
    llava-v1.5-7b & 76.56 (1.14) & 20.48 (0.24) & 21.14 (0.27) & 22.06 (0.31) & 30.88 (0.12) & 31.77 (0.11) & 32.81 (0.17) \\
    llava-v1.5-13b & 73.37 (1.34) & 20.50 (0.37) & 21.08 (0.29) & 21.85 (0.25) & 30.95 (0.12) & 31.83 (0.12) & 32.67 (0.14) \\
    internlm-xcomposer2-vl-7b & 78.60 (1.46) & 21.16 (0.31) & 20.60 (0.24) & 21.36 (0.32) & 31.60 (0.06) & 31.23 (0.09) & 32.02 (0.12) \\
    Muffin-13B & 70.88 (1.60) & 20.67 (0.24) & 20.77 (0.29) & 21.07 (0.33) & 30.71 (0.10) & 30.73 (0.07) & 30.96 (0.07) \\
    RLHF-V & 76.37 (1.48) & 21.52 (0.19) & 21.57 (0.23) & 21.78 (0.25) & 31.47 (0.13) & 31.42 (0.12) & 31.72 (0.07) \\
    llava-v1.6-vicuna-7b & 79.61 (1.85) & 21.26 (0.27) & 21.39 (0.30) & 22.06 (0.38) & 31.92 (0.17) & 32.11 (0.08) & 32.66 (0.13) \\
    llava-v1.6-vicuna-13b & 78.97 (1.44) & 21.29 (0.25) & 21.40 (0.28) & 21.97 (0.34) & 31.95 (0.13) & 32.16 (0.10) & 32.70 (0.12) \\
    Qwen-VL & 65.73 (0.51) & 15.58 (0.27) & 15.04 (0.20) & 15.83 (0.18) & 24.23 (0.32) & 23.24 (0.32) & 24.23 (0.32) \\
    \hline
    Pearson's r (p value) & \textcolor{gray}{100.00 (0.00)} & 75.81 (0.01)* & 69.31 (0.03)* & 72.30 (0.02)* & 77.62 (0.01)* & 72.32 (0.02)* & 74.36 (0.01)* \\
    \bottomrule

    \toprule
    \multirow{2}{*}{Model} & OIv7 POPE & \multicolumn{6}{c}{Pfram($\mathcal{F}; \mathcal{G}_{obj}| \phi_{\mathrm{kNN}},$ OIv7) (std.)} \\
    & acc (std.) & k=200 input & k=200 mean & k=200 max & k=400 input & k=400 mean & k=400 max \\
    \hline
    instructblip-vicuna-7b & 73.21 (1.29) & 39.29 (0.20) & 38.84 (0.06) & 39.29 (0.20) & 52.02 (0.74) & 51.25 (0.66) & 52.02 (0.74) \\
    instructblip-vicuna-13b & 67.54 (1.41) & 38.23 (0.04) & 38.74 (0.04) & 39.12 (0.10) & 50.71 (0.63) & 51.16 (0.63) & 51.49 (0.55) \\
    llava-v1.5-7b & 76.56 (1.14) & 39.00 (0.18) & 40.21 (0.19) & 41.45 (0.20) & 51.46 (0.48) & 52.84 (0.53) & 54.15 (0.54) \\
    llava-v1.5-13b & 73.37 (1.34) & 39.07 (0.24) & 40.30 (0.23) & 41.34 (0.22) & 51.47 (0.51) & 52.96 (0.50) & 54.16 (0.49) \\
    internlm-xcomposer2-vl-7b & 78.60 (1.46) & 39.88 (0.13) & 39.48 (0.07) & 40.30 (0.15) & 51.84 (0.45) & 51.79 (0.55) & 52.95 (0.53) \\
    Muffin-13B & 70.88 (1.60) & 38.64 (0.06) & 38.57 (0.05) & 38.80 (0.05) & 50.94 (0.65) & 50.70 (0.64) & 51.14 (0.64) \\
    RLHF-V & 76.37 (1.48) & 39.48 (0.14) & 39.28 (0.08) & 39.71 (0.09) & 51.75 (0.54) & 51.26 (0.53) & 51.98 (0.53) \\
    llava-v1.6-vicuna-7b & 79.61 (1.85) & 40.36 (0.23) & 40.59 (0.21) & 41.12 (0.20) & 53.02 (0.49) & 53.21 (0.53) & 53.92 (0.51) \\
    llava-v1.6-vicuna-13b & 78.97 (1.44) & 40.35 (0.21) & 40.59 (0.19) & 41.24 (0.23) & 53.00 (0.52) & 53.27 (0.53) & 54.06 (0.49) \\
    Qwen-VL & 65.73 (0.51) & 32.16 (0.27) & 31.01 (0.34) & 32.16 (0.27) & 44.34 (0.83) & 43.08 (0.90) & 44.34 (0.83) \\
    \hline
    Pearson's r (p value) & \textcolor{gray}{100.00 (0.00)} & 79.37 (0.01)* & 74.24 (0.01)* & 76.31 (0.01)* & 78.01 (0.01)* & 74.09 (0.01)* & 77.22 (0.01)* \\
    \bottomrule
    \end{tabular}}
    \caption{Results of OIv7 POPE acc and Pfram($\mathcal{F}; \mathcal{G}_{obj}| \phi_{\mathrm{kNN}},$ OIv7) scores. In the last row we report the Pearson's r correlation (and p value) between each column and OIv7 POPE acc. *: Statistically significant with $p \le 0.05$.} \label{tab:OIv7_SOR_kNN}
\end{table}

\begin{table}[t]
    \centering
    \resizebox{\textwidth}{!}{
    \begin{tabular}{c|c|c|c|c|c|c|c}
    \toprule
    \multirow{2}{*}{Model} & OIv7 POPE & \multicolumn{6}{c}{Pfram($\mathcal{F}; \mathcal{G}_{des}| \phi_{\mathrm{kNN}},$ OIv7) (std.)} \\
    & acc & k=25 input & k=25 mean & k=25 max & k=100 input & k=100 mean & k=100 max \\
    \hline
    instructblip-vicuna-7b & 73.21 (1.29) & 36.93 (0.52) & 37.40 (0.46) & 38.15 (0.49) & 44.72 (0.34) & 45.11 (0.24) & 45.56 (0.22) \\
    instructblip-vicuna-13b & 67.54 (1.41) & 35.70 (0.37) & 36.68 (0.44) & 37.43 (0.59) & 43.74 (0.24) & 44.50 (0.36) & 44.94 (0.47) \\
    llava-v1.5-7b & 76.56 (1.14) & 34.36 (0.15) & 33.87 (0.23) & 35.07 (0.20) & 42.30 (0.29) & 41.85 (0.28) & 43.16 (0.28) \\
    llava-v1.5-13b & 73.37 (1.34) & 34.32 (0.26) & 33.52 (0.20) & 34.94 (0.17) & 42.23 (0.30) & 41.57 (0.25) & 43.13 (0.29) \\
    internlm-xcomposer2-vl-7b & 78.60 (1.46) & 34.37 (0.17) & 34.80 (0.24) & 37.15 (0.35) & 42.75 (0.21) & 43.34 (0.17) & 45.41 (0.20) \\
    Muffin-13B & 70.88 (1.60) & 38.78 (0.30) & 39.17 (0.30) & 39.59 (0.35) & 46.67 (0.19) & 46.94 (0.19) & 47.29 (0.22) \\
    RLHF-V & 76.37 (1.48) & 38.20 (0.32) & 38.65 (0.24) & 39.01 (0.18) & 45.59 (0.21) & 45.85 (0.25) & 46.46 (0.20) \\
    llava-v1.6-vicuna-7b & 79.61 (1.85) & 33.38 (0.21) & 34.20 (0.25) & 35.55 (0.33) & 41.74 (0.16) & 42.46 (0.20) & 43.25 (0.24) \\
    llava-v1.6-vicuna-13b & 78.97 (1.44) & 33.39 (0.31) & 33.97 (0.26) & 34.97 (0.25) & 41.57 (0.17) & 42.06 (0.22) & 42.61 (0.28) \\
    Qwen-VL & 65.73 (0.51) & 25.22 (0.60) & 24.55 (0.54) & 25.99 (0.52) & 32.52 (0.57) & 31.39 (0.53) & 32.52 (0.57) \\
    \hline
    Pearson's r (p value) & \textcolor{gray}{100.00 (0.00)} & 30.40 (0.39) & 31.86 (0.37) & 38.21 (0.28) & 35.14 (0.32) & 38.03 (0.28) & 42.93 (0.22) \\
    \bottomrule

    \toprule
    \multirow{2}{*}{Model} & OIv7 POPE & \multicolumn{6}{c}{Pfram($\mathcal{F}; \mathcal{G}_{des}| \phi_{\mathrm{kNN}},$ OIv7) (std.)} \\
    & acc & k=200 input & k=200 mean & k=200 max & k=400 input & k=400 mean & k=400 max \\
    \hline
    instructblip-vicuna-7b & 73.21 (1.29) & 50.35 (0.53) & 50.75 (0.42) & 51.09 (0.41) & 57.42 (0.55) & 57.80 (0.41) & 58.17 (0.37) \\
    instructblip-vicuna-13b & 67.54 (1.41) & 49.66 (0.40) & 50.12 (0.41) & 50.37 (0.37) & 57.15 (0.43) & 57.23 (0.42) & 57.65 (0.39) \\
    llava-v1.5-7b & 76.56 (1.14) & 48.51 (0.38) & 48.25 (0.36) & 49.38 (0.37) & 56.13 (0.45) & 56.17 (0.48) & 57.05 (0.43) \\
    llava-v1.5-13b & 73.37 (1.34) & 48.39 (0.39) & 48.02 (0.33) & 49.43 (0.36) & 56.19 (0.42) & 56.06 (0.45) & 57.16 (0.42) \\
    internlm-xcomposer2-vl-7b & 78.60 (1.46) & 49.14 (0.31) & 49.54 (0.31) & 51.32 (0.36) & 56.88 (0.43) & 57.08 (0.43) & 58.73 (0.48) \\
    Muffin-13B & 70.88 (1.60) & 52.18 (0.42) & 52.34 (0.43) & 52.60 (0.42) & 59.05 (0.53) & 58.92 (0.55) & 59.28 (0.53) \\
    RLHF-V & 76.37 (1.48) & 51.10 (0.41) & 51.13 (0.46) & 51.96 (0.40) & 58.03 (0.52) & 57.54 (0.55) & 58.63 (0.52) \\
    llava-v1.6-vicuna-7b & 79.61 (1.85) & 48.41 (0.25) & 48.92 (0.30) & 49.36 (0.32) & 56.67 (0.45) & 56.82 (0.45) & 57.13 (0.44) \\
    llava-v1.6-vicuna-13b & 78.97 (1.44) & 48.15 (0.26) & 48.47 (0.31) & 48.81 (0.33) & 56.40 (0.43) & 56.44 (0.44) & 56.78 (0.40) \\
    Qwen-VL & 65.73 (0.51) & 38.73 (0.75) & 37.59 (0.72) & 38.73 (0.75) & 47.18 (0.87) & 46.27 (0.86) & 47.18 (0.87) \\
    \hline
    Pearson's r (p value) & \textcolor{gray}{100.00 (0.00)} & 42.21 (0.22) & 44.42 (0.20) & 48.30 (0.16) & 49.02 (0.15) & 50.58 (0.14) & 53.35 (0.11) \\
    \bottomrule
    \end{tabular}}
    \caption{Results of OIv7 POPE acc and Pfram($\mathcal{F}; \mathcal{G}_{des}| \phi_{\mathrm{kNN}},$ OIv7) scores. In the last row we report the Pearson's r correlation (and p value) between each column and OIv7 POPE acc.} \label{tab:OIv7_SC_kNN}
\end{table}

\begin{table}[t]
    \centering
    \resizebox{\textwidth}{!}{
    \begin{tabular}{c|c|c|c|c|c|c|c}
    \toprule
    \multirow{2}{*}{Model} & AMBER POPE & \multicolumn{6}{c}{Pfram($\mathcal{F}; \mathcal{G}_{obj}| \phi_{\mathrm{kNN}},$ AMBER)} \\
    & acc & k=25 input & k=25 mean & k=25 max & k=100 input & k=100 mean & k=100 max \\
    \hline
    instructblip-vicuna-7b & 82.47 & 35.49 & 35.92 & 36.60 & 47.92 & 48.58 & 49.46 \\
    instructblip-vicuna-13b & 76.87 & 34.90 & 35.09 & 35.74 & 47.13 & 47.42 & 48.03 \\
    llava-v1.5-7b & 76.50 & 34.81 & 36.17 & 38.88 & 48.10 & 50.94 & 54.93 \\
    llava-v1.5-13b & 80.33 & 34.72 & 36.35 & 38.59 & 48.37 & 51.92 & 55.29 \\
    internlm-xcomposer2-vl-7b & 84.00 & 34.79 & 35.01 & 36.98 & 49.62 & 50.34 & 53.55 \\
    Muffin-13B & 77.13 & 34.84 & 34.84 & 35.25 & 47.12 & 46.93 & 47.27 \\
    RLHF-V & 81.17 & 36.76 & 36.44 & 36.76 & 47.84 & 47.11 & 47.84 \\
    llava-v1.6-vicuna-7b & 81.97 & 35.60 & 36.40 & 38.29 & 51.08 & 52.25 & 54.36 \\
    llava-v1.6-vicuna-13b & 83.50 & 35.17 & 36.50 & 38.25 & 50.93 & 52.86 & 54.70 \\
    Qwen-VL & 74.93 & 25.88 & 25.99 & 28.14 & 35.53 & 34.95 & 36.16 \\
    \hline
    Pearson's r (p value) & \textcolor{gray}{100.00 (0.00)} & 57.56 (0.08) & 57.86 (0.08) & 57.29 (0.08) & 69.78 (0.02)* & 64.02 (0.05)* & 60.20 (0.07) \\
    \bottomrule

    \toprule
    \multirow{2}{*}{Model} & AMBER POPE & \multicolumn{6}{c}{Pfram($\mathcal{F}; \mathcal{G}_{obj}| \phi_{\mathrm{kNN}},$ AMBER)} \\
    & acc & k=200 input & k=200 mean & k=200 max & k=400 input & k=400 mean & k=400 max \\
    \hline
    instructblip-vicuna-7b & 82.47 & 58.19 & 58.57 & 59.16 & 77.47 & 77.52 & 77.76 \\
    instructblip-vicuna-13b & 76.87 & 58.03 & 57.87 & 58.29 & 76.99 & 77.30 & 77.71 \\
    llava-v1.5-7b & 76.50 & 59.36 & 62.30 & 65.75 & 77.93 & 79.19 & 80.46 \\
    llava-v1.5-13b & 80.33 & 59.56 & 63.23 & 66.48 & 77.77 & 79.51 & 80.81 \\
    internlm-xcomposer2-vl-7b & 84.00 & 59.55 & 60.70 & 63.58 & 76.21 & 77.40 & 80.30 \\
    Muffin-13B & 77.13 & 57.22 & 57.00 & 57.26 & 76.25 & 76.43 & 76.72 \\
    RLHF-V & 81.17 & 57.68 & 56.67 & 57.79 & 76.15 & 76.00 & 76.32 \\
    llava-v1.6-vicuna-7b & 81.97 & 62.47 & 63.46 & 65.10 & 79.04 & 79.66 & 80.44 \\
    llava-v1.6-vicuna-13b & 83.50 & 62.32 & 63.94 & 65.48 & 78.83 & 80.07 & 81.02 \\
    Qwen-VL & 74.93 & 47.12 & 46.44 & 47.12 & 68.51 & 68.18 & 68.51 \\
    \hline
    Pearson's r (p value) & \textcolor{gray}{100.00 (0.00)} & 66.61 (0.04)* & 60.11 (0.07) & 57.04 (0.09) & 55.88 (0.09) & 55.77 (0.09) & 60.38 (0.06) \\
    \bottomrule
    \end{tabular}}
    \caption{Results of AMBER POPE acc and Pfram($\mathcal{F}; \mathcal{G}_{obj}| \phi_{\mathrm{kNN}},$ AMBER) scores. In the last row we report the Pearson's r correlation (and p value) between each column and AMBER POPE acc. *: Statistically significant with $p \le 0.05$.} \label{tab:AMBER_SOR_kNN}
\end{table}

\begin{table}[t]
    \centering
    \resizebox{\textwidth}{!}{
    \begin{tabular}{c|c|c|c|c|c|c|c}
    \toprule
    \multirow{2}{*}{Model} & AMBER POPE & \multicolumn{6}{c}{Pfram($\mathcal{F}; \mathcal{G}_{des}| \phi_{\mathrm{kNN}},$ AMBER)} \\
    & acc & k=25 input & k=25 mean & k=25 max & k=100 input & k=100 mean & k=100 max \\
    \hline
    instructblip-vicuna-7b & 82.47 & 52.31 & 52.66 & 53.37 & 58.04 & 58.35 & 58.96 \\
    instructblip-vicuna-13b & 76.87 & 52.13 & 51.83 & 52.30 & 58.09 & 57.94 & 58.27 \\
    llava-v1.5-7b & 76.50 & 45.79 & 44.93 & 46.80 & 53.60 & 53.12 & 54.74 \\
    llava-v1.5-13b & 80.33 & 45.49 & 43.84 & 46.41 & 53.28 & 52.51 & 54.61 \\
    internlm-xcomposer2-vl-7b & 84.00 & 42.03 & 43.33 & 46.90 & 50.21 & 51.44 & 54.95 \\
    Muffin-13B & 77.13 & 52.38 & 52.66 & 53.10 & 58.75 & 58.78 & 59.06 \\
    RLHF-V & 81.17 & 52.07 & 52.19 & 52.48 & 57.11 & 56.50 & 57.55 \\
    llava-v1.6-vicuna-7b & 81.97 & 44.65 & 44.89 & 46.32 & 53.48 & 53.26 & 54.05 \\
    llava-v1.6-vicuna-13b & 83.50 & 44.00 & 44.24 & 45.26 & 52.99 & 52.80 & 53.64 \\
    Qwen-VL & 74.93 & 33.63 & 33.86 & 36.43 & 38.34 & 37.61 & 38.50 \\
    \hline
    Pearson's r (p value) & \textcolor{gray}{100.00 (0.00)} & 13.24 (0.72) & 18.41 (0.61) & 23.22 (0.52) & 26.93 (0.45) & 31.47 (0.38) & 38.71 (0.27) \\
    \bottomrule

    \toprule
    \multirow{2}{*}{Model} & AMBER POPE & \multicolumn{6}{c}{Pfram($\mathcal{F}; \mathcal{G}_{des}| \phi_{\mathrm{kNN}},$ AMBER)} \\
    & acc & k=200 input & k=200 mean & k=200 max & k=400 input & k=400 mean & k=400 max \\
    \hline
    instructblip-vicuna-7b & 82.47 & 64.59 & 64.74 & 65.29 & 76.99 & 77.06 & 77.54 \\
    instructblip-vicuna-13b & 76.87 & 64.82 & 64.43 & 64.88 & 77.46 & 77.24 & 77.53 \\
    llava-v1.5-7b & 76.50 & 61.58 & 61.41 & 62.66 & 76.56 & 76.73 & 77.08 \\
    llava-v1.5-13b & 80.33 & 61.27 & 60.85 & 62.57 & 76.27 & 76.44 & 76.88 \\
    internlm-xcomposer2-vl-7b & 84.00 & 58.52 & 59.53 & 62.34 & 74.37 & 74.91 & 76.57 \\
    Muffin-13B & 77.13 & 65.18 & 65.09 & 65.40 & 77.57 & 77.42 & 77.66 \\
    RLHF-V & 81.17 & 63.73 & 62.71 & 64.12 & 76.78 & 76.02 & 77.07 \\
    llava-v1.6-vicuna-7b & 81.97 & 61.77 & 61.34 & 62.27 & 76.78 & 76.69 & 77.02 \\
    llava-v1.6-vicuna-13b & 83.50 & 61.32 & 60.98 & 61.84 & 76.43 & 76.51 & 76.70 \\
    Qwen-VL & 74.93 & 48.21 & 47.21 & 48.21 & 68.94 & 68.65 & 69.00 \\
    \hline
    Pearson's r (p value) & \textcolor{gray}{100.00 (0.00)} & 30.02 (0.40) & 34.31 (0.33) & 41.42 (0.23) & 32.51 (0.36) & 36.81 (0.30) & 45.49 (0.19) \\
    \bottomrule
    \end{tabular}}
    \caption{Results of AMBER POPE acc and Pfram($\mathcal{F}; \mathcal{G}_{des}| \phi_{\mathrm{kNN}},$ AMBER) scores. In the last row we report the Pearson's r correlation (and p value) between each column and AMBER POPE acc.} \label{tab:AMBER_SC_kNN}
\end{table}

\end{document}